%% file: main.tex
\documentclass{article}

\PassOptionsToPackage{numbers}{natbib}
\usepackage[preprint]{neurips_2026}
\usepackage[T1]{fontenc}
\usepackage[utf8]{inputenc}
\usepackage{hyperref}
\usepackage{url}
\usepackage{booktabs}
\usepackage{array}
\usepackage{multirow}
\usepackage{amsmath}
\usepackage{amsfonts}
\usepackage{nicefrac}
\usepackage{microtype}
\usepackage{xcolor}
\usepackage{graphicx}
\usepackage{subcaption}
\usepackage{float}
\usepackage{enumitem}

\newcommand{\tightparagraph}[1]{\vspace{0.1ex}\noindent\textbf{#1}}

\title{Training-Free Temporal Abstraction for \\General Video Understanding}
\author{%
  Etienne Casanova \\
  California Institute of Technology \\
  \And
  Sevan Brodjian \\
  California Institute of Technology \\
  \And
  Pietro Perona \\
  California Institute of Technology
}

\date{}
\begin{document}
\maketitle

% general ideaa: modest - we propose a method of reasoning about video that  - video-text foundation approach has been overlooked
% Stitch method is a way of temporal abstraction

\begin{abstract}
Videos are expensive to analyze frame by frame, yet many video understanding tasks depend on knowing where relevant moments occur. A system may need to find when an action changes, locate the segment described by a sentence, or choose a few frames for a vision-language model. Existing methods often solve these problems separately, using task-specific training data or specialized architectures. We study whether a pretrained video-text model can provide enough temporal structure to support several of these tasks at once. We present \textbf{STITCH}, a training-free method that divides a video into semantically meaningful temporal chunks. STITCH embeds short video windows with a frozen video-text backbone and detects changes in the resulting embedding sequence. These chunks are computed once per video and reused across tasks. We evaluate STITCH on generic event boundary detection, language-based moment retrieval, and frame selection for long-video VLM reasoning. Across all three settings, STITCH remains competitive with more specialized methods while requiring no task-specific training, with especially clear gains when only a small number of frames or tokens can be processed. These results suggest that reusable temporal abstraction is a promising direction for general video understanding, allowing dense video streams to be converted once into semantic units that can be localized, retrieved, sampled, or reasoned over by downstream systems.

% \red{\textbf{TODO: Checklist (last pages)}}

% Long videos are difficult to process efficiently because relevant events are sparse and unevenly distributed in time. We present STITCH, a training-free method for semantic temporal chunking that turns a pretrained video-text model into a reusable temporal abstraction layer. STITCH embeds short video windows with InternVideo2, detects changes in embedding space, and partitions each video into adaptive chunks, with optional lightweight post-processing such as chunk merging. The same chunking procedure is then reused across generic event boundary detection, language-based moment retrieval, and frame selection for long-video VLM reasoning. Across these settings, STITCH is competitive with more specialized methods and is especially useful under constrained frame or token budgets. This suggests that pretrained video-text embeddings contain enough temporal structure to support simple training-free temporal abstraction for video understanding.
\end{abstract}
% Instead of task-specific models: Traditionally, this is achieved with a task-specific models, however these models don't generalize to other tasks.
% notes: make it more understandable, especially about specific video tasks - what does that mean - make more clear
% reusable instead of adaptive - or just remove adaptive. remove adaptive

% Make sure fig 1 is on pg 2
\begin{figure}[t]
    \centering
    \includegraphics[width=0.9\linewidth]{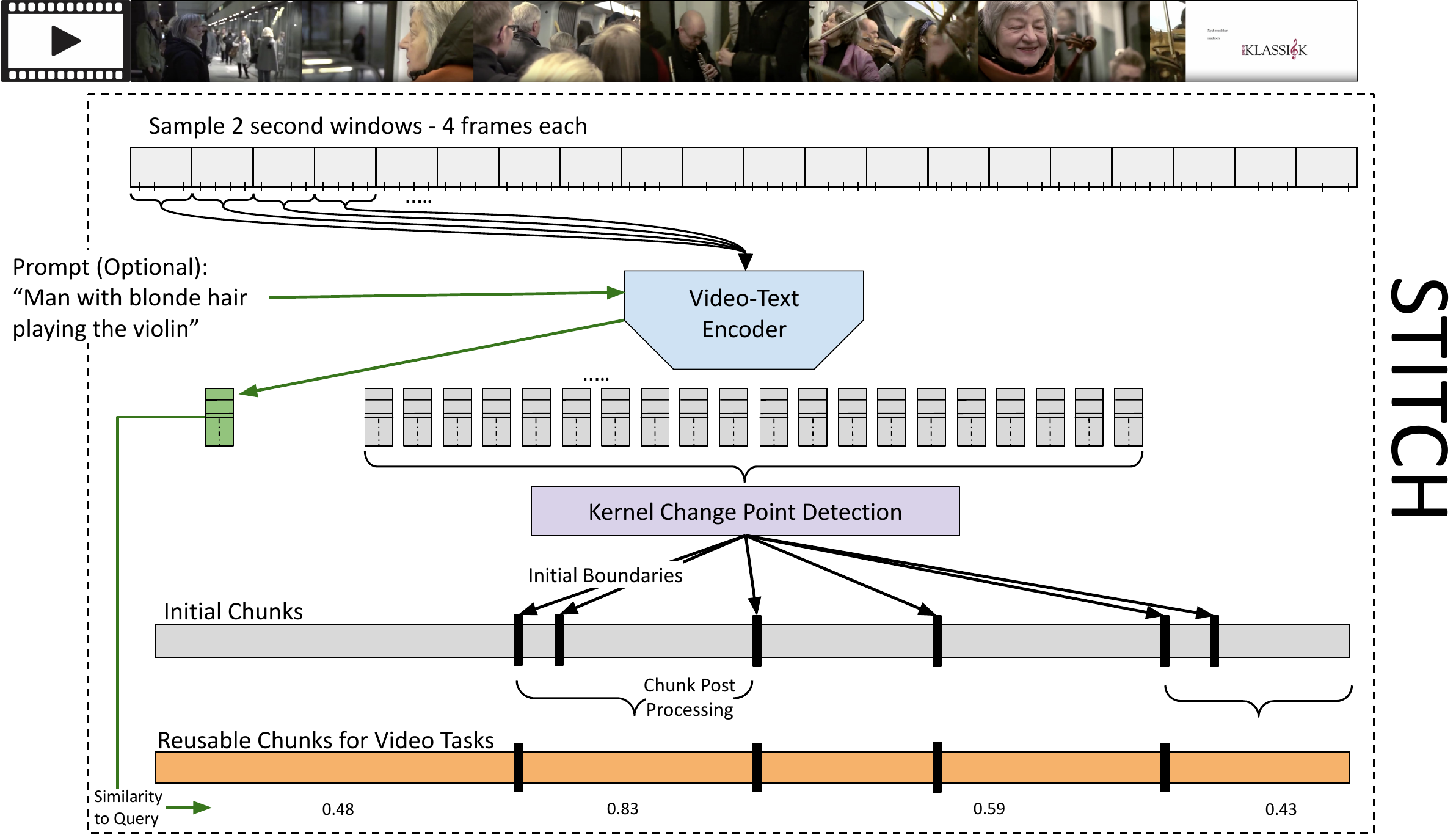}
    \caption{\textbf{STITCH converts semantic change into a reusable video timeline.} As described in Section~\ref{sec:approach}, STITCH first encodes short video windows with a frozen video-text backbone, semantic change points are then detected in embedding space, and the video is partitioned into adaptive chunks. The same chunked timeline can be interpreted directly as event boundaries, scored with a language query for retrieval, or used to select compact visual evidence for long-video reasoning.}
    \label{fig:overview}
    \vspace{-1em}
\end{figure}

\input{sections/1_intro}
\input{sections/2_related}
\input{sections/3_approach}
% TODO: 
% This should be the generalized chunking approach
% We include figure here, we include CPD, leave the specifics of how we can use this for later
% The mathmatical formula / foundation should be here and should be easily transferrable to the rest of the paper
\input{sections/4_experiments}
% TODO: we want to note that we chose datasets and tasks in order to get a variety of tasks, to show that this architecture can be applied to general video tasks
\input{sections/5_app1_gebd}
% Event Detection: Tapos, Kinetics GEBD
\input{sections/6_app2_mr}
% Moment Retrieval: qv highlights, activitynetqa 
\input{sections/7_app3_vlm}
\input{full_page_methods_fig}
\input{main_results_table}
% VLM QA: VideoMME, LongVideoBench, MVLU, activitynetQA (if time), LVBench (if time aswell)
% Potentially bar plot - uniform vs ours - baseline of no frames gives (so straight up passing to the VLM), reason is so that VLM might just be able to 
% Also want to note that we are not interested in things like prompts and adding metadata and subtitles, as that will further noise our method

\input{sections/8_discussion}
% TODO: talk about how the system is real time

\section{Conclusion}
\label{sec:conclusion}
Dense video does not need to be processed only as frames, fixed clips, or task-specific proposals. STITCH shows that a frozen video-text model can provide a reusable temporal structure by converting short-window embeddings into semantic chunks. The same chunked timeline supports event boundary detection, language-based moment retrieval, and frame selection for long-video VLM reasoning without training task-specific heads. Across these settings, STITCH is strongest against training-free, zero-shot, and unsupervised methods, while remaining competitive with many supervised systems that optimize directly for a target benchmark.

These results support temporal abstraction as a practical interface between raw video and downstream reasoning. Once a video is converted into semantic units, those units can be localized, scored, sampled, or stored according to the needs of the task. This separates the problem of finding useful temporal structure from the problem of using that structure for a particular application. As video models become stronger and long-context systems become more common, reusable temporal timelines may become an important front end for general video understanding.

% TODO: Proof of principal for this direction that people haven't gone for before. Worth exploring further

% Future work: Could discuss video to video and other retrieval tasks that we believe that this system could generalize very well to. In addition, testing on more powerful models - since we do not use the best version of internvideo2 (there is a larger, more powerful version with more params that we could not run because of compute)
% For future work it's also worth highlighting how the real-time processing speeds on consumer devices make this a promising setup for streaming-based video setups (e.g. robotics or spatial computing). Furthermore, since the chunk embeddings only need to be calculated once, this offers a compressed sequence to store and operate over instead of raw video buffers for long-context tasks.

% Limitations: Sequence of events we would not be good at understandings

% Mention: We looked into agentic video understanding

\bibliographystyle{plainnat}
\bibliography{main}

\appendix
\include{supplemental.tex}

\end{document}

%% file: sections/1_intro.tex
\section{Introduction}
\label{sec:intro}

Video understanding often begins by turning a dense stream of frames into a smaller set of meaningful temporal units. A boundary detector must identify when one event gives way to another. A moment retrieval system must find the part of a video that matches a sentence. A video-language model may need to choose which frames to inspect when the full video is too large to process densely. Although these problems are usually studied separately, they share a common difficulty: the system must identify when meaningful information occurs. A useful video representation should therefore adapt its temporal units to the content, rather than relying only on individual frames or fixed-length clips.

Current video systems often solve temporal localization by building machinery for one task at a time. Supervised boundary detectors learn to predict event changes, temporal grounding models learn to match language queries to video spans, and frame selection methods learn or design rules for choosing visual evidence for video reasoning. These approaches can be highly effective when training data is available for the target setting, but their temporal representations are often tied to the task for which they were built. A more reusable system would separate the problem of finding useful temporal units from the problem of applying those units to a particular downstream task.

Pretrained video-text models make this separation more plausible. These models map short video clips and text into a shared representation space, where semantically related visual and linguistic content should be close together \citep{radford2021clip, ma2022xclip, Wang2024InternVideo2SF}. If nearby video windows have similar representations, they may belong to the same event or semantic period. If the representation changes sharply, the video may have crossed into a new event, action, or scene. The embedding stream of a frozen video-text model may therefore be sufficient to define useful temporal units for downstream video tasks.

We introduce \textbf{STITCH} (\textbf{S}emantic \textbf{T}emporal \textbf{I}nference-\textbf{T}ime \textbf{CH}unking), a training-free method for dividing videos into adaptive semantic chunks. Given a video, STITCH samples short temporal windows, embeds each window with InternVideo2 \citep{Wang2024InternVideo2SF}, and detects changes in the resulting sequence of embeddings. These changes define chunks whose lengths adapt to the content. The chunks are computed without labels, without task-specific training, and without knowing the downstream query in advance. Once computed, they provide a reusable temporal structure for understanding the video.

STITCH tests temporal abstraction as a shared interface between dense video encoders and downstream video understanding systems. Prior work on event boundary detection, temporal grounding, and long-video frame selection has produced strong specialized methods, often developed within separate literatures. STITCH extracts a set of query-independent chunks from a frozen video-text backbone, then reuses that temporal structure for localization, retrieval, and frame-budgeted reasoning. The goal is not to replace supervised systems when task-specific labels are available. It is to show that pretrained video-text embeddings already contain temporal organization that can be extracted once, reused across tasks, and evaluated as a common abstraction layer. Across all three settings, STITCH remains competitive with more specialized approaches while requiring no task-specific training, with the clearest gains appearing when long-video VLMs must reason from a small visual budget.

Our main contributions are as follows:
\begin{enumerate}[
    label=(\arabic*),
    leftmargin=*,
    align=left,
    itemsep=0.25em,
    topsep=0pt,
    parsep=0pt,
    partopsep=0pt
]
\item \textbf{Training-free semantic temporal chunking.} We introduce STITCH, which uses a frozen video-text backbone, change-point detection, and lightweight post-processing to divide videos into adaptive semantic chunks without labels or task-specific training.
\item \textbf{One reusable video timeline across tasks.} We show that the same query-independent chunk structure can support generic event boundary detection, language-based moment retrieval, and frame selection for long-video VLM reasoning.
% \item \textbf{Evidence for temporal abstraction under limited budgets.} We evaluate STITCH across localization, retrieval, and reasoning benchmarks, showing that a simple training-free front end can remain competitive with specialized methods and is especially useful when downstream models cannot process video densely.
\item \textbf{Temporal abstraction under limited budgets.} We evaluate STITCH across localization, retrieval, and reasoning benchmarks, showing that a simple training-free front end can remain competitive with specialized methods, particularly when downstream systems cannot process video densely.
\end{enumerate}

%% file: sections/2_related.tex
\section{Related Work}
\label{sec:prior}

\paragraph{Adaptive temporal segmentation and chunking.}
A long video is rarely well represented by a fixed temporal grid. This has motivated work on event segmentation, scene boundary detection, action segmentation, and adaptive tokenization, where dense frame processing is replaced by temporally coherent units. The closest prior work to our setting is KTS Revisited, which uses an unsupervised KTS tokenizer for adaptive sampling in long-form video classification and temporal action localization \citep{poursaeed2023revisitingkerneltemporal}. Its segmentation step is training-free, but its reported gains still come from feeding KTS-sampled inputs into trained downstream task models. STITCH instead keeps the full pipeline training-free: a frozen video-text backbone defines the chunks, and the same embedding stream supports boundary prediction, text-based scoring, and frame selection without fitting task-specific heads. Unlike STITCH, many competitive event-boundary methods train explicit boundary predictors or generative boundary models specialized for that setting \citep{tan2023temporalperceiver,zheng2024dybdet,hwang2025diffgebd}.

\paragraph{Moment retrieval and training-free temporal grounding.}
Language-based moment retrieval is typically approached with supervised models that learn query-conditioned temporal representations or proposal scoring functions \citep{krishna2017densecaption,Lei2021QVHighlightsDM,luo2020learning2dtemporal,yuan2019scdm,Moon_2023_CVPR,Sun_2024_AAAI_TRDETR}. Recent training-free approaches instead use pretrained vision-language or language models to generate, score, refine, or filter candidate spans at inference time \citep{Luo2024FrozenVLMVMR,zheng2024tfvtg,xu2025zeroshot_vmr_mllm,qu2024chatvtg}. TAG reduces semantic fragmentation with temporal pooling, temporal-coherence clustering, and similarity adjustment \citep{Lee_2025_BMVC}, while Point-to-Span targets hour-long videos with adaptive span generation and query decomposition \citep{Jeon2025PointTS}. STITCH shares the training-free motivation of these methods, but differs in where temporal structure enters the pipeline. Rather than constructing query-dependent proposals, captions, or refined search spaces for each query, STITCH first computes a query-independent chunking of the video and then scores those reusable chunks against text. This makes the video-side computation reusable across queries and lets the same temporal units serve tasks beyond moment retrieval.

\paragraph{Long-video VLM frame selection and compression.}
Long-video VLMs have limited frame and token budgets, making temporal selection a central design problem. Uniform sampling is simple but can miss short or sparse evidence~\citep{Chasmai2025MomentSampling}. Recent methods address this limitation through adaptive keyframe selection, learned frame querying, redundancy reduction, narrative threading, or training-free selection heuristics \citep{yu2025framevoyager,liu2025bolt,tang2025adaptive,fang2026threading,li2026frameoracle}. Closely related methods use training-free or semantic-boundary-aware selection: VideoTree builds a query-adaptive hierarchy \citep{Wang2024VideoTree}, while WFS-SB detects semantic boundaries with wavelet analysis and samples around narrative shifts \citep{chen2026wfsb}. STITCH also uses non-uniform selection, but guides it with reusable chunks computed once from InternVideo2 embeddings \citep{Wang2024InternVideo2SF}.

\paragraph{Unified localization frameworks.}
Several recent systems argue against treating temporal localization tasks as separate problems. UniVTG unifies video-language temporal grounding formats through supervised pretraining over diverse temporal labels \citep{Lin2023UniVTG}. UnLoc unifies moment retrieval, temporal action localization, and action segmentation with a CLIP-based localization architecture \citep{Yan2023UnLoc}, and TimeLoc frames temporal action localization, temporal video grounding, moment retrieval, and GEBD as timestamp localization within an end-to-end trained model \citep{Zhang2025TimeLoc}. These frameworks each unify related localization or grounding tasks by learning models that predict task-specific temporal outputs. Their benchmark suites therefore overlap with ours only partially, and their notion of unification remains centered on localization-style prediction. STITCH instead makes the chunking itself reusable, with one frozen, query-independent partition used for event boundaries, text retrieval, and frame-budgeted VLM reasoning. In this sense, STITCH extends the unified-localization perspective from learned timestamp prediction to reusable temporal abstraction.

% \paragraph{Positioning.}
% Taken together, prior work provides strong evidence for each component of our motivation: adaptive temporal units are useful, frozen video-language models can support training-free grounding, unified localization is desirable, and long-video VLMs need better frame selection. STITCH sits at the intersection of these lines rather than claiming novelty in any single ingredient. Its central claim is more concrete: a single training-free chunking procedure, built on one frozen video-text backbone and computed once per video, can serve as a reusable temporal structure across GEBD, moment retrieval, and VLM frame selection. This makes STITCH a general-method baseline for temporal abstraction in video understanding, complementary to both specialized supervised systems and query-specific training-free pipelines.

%% file: sections/3_approach.tex
\section{Approach}
\label{sec:approach}

STITCH is a training-free temporal abstraction pipeline that computes a reusable chunk structure for a video. Given a video, we encode short temporal windows with one frozen video-text backbone, detect changes in the resulting embedding sequence, and partition the video into adaptive chunks. The chunks are computed once, independently of any downstream task, and can then be reused for event boundary detection, language-based moment retrieval, or frame selection for long-video reasoning. This design keeps the video-side computation fixed while allowing different lightweight readouts to be applied on top.
We include an anonymized implementation with the submission at \url{https://anonymous.4open.science/r/STITCH-video-understanding-43D6}.

\subsection{Dense window embeddings}
\label{subsec:dense_embeddings}

STITCH begins by converting the video into a time-indexed stream of semantic embeddings. We sample the video as a sequence of short temporal windows
\[
W = \{w_1, w_2, \dots, w_T\},
\]
where each window covers a fixed duration span (typically 0.5--2 seconds depending on the task) and adjacent windows are non-overlapping. Each window is encoded from a small number of uniformly sampled frames (typically 4) using the frozen video-text backbone,
\[
\mathbf{e}_t = f_{\mathrm{vid}}(w_t) \in \mathbb{R}^d,
\qquad t = 1,\dots,T.
\]
In all experiments, we use the same frozen InternVideo2-Stage2-1B encoder for every task \citep{Wang2024InternVideo2SF}, the smaller of the two released InternVideo2 encoder models. The embeddings are $\ell_2$-normalized and treated as a temporal signal over the video.

This shared embedding stream is the common representation used throughout the method. If nearby windows have similar embeddings, they likely belong to the same event, action, or semantic period. If the embedding changes sharply, the video may have crossed into a new temporal unit. Because InternVideo2 maps video and text into a joint embedding space, the same window embeddings can later be used both for chunking and for query-based scoring.

\subsection{Semantic temporal chunking}
\label{subsec:chunking}

STITCH forms chunks by detecting semantic changes in the embedding sequence. A direct signal is the cosine similarity between adjacent windows: low adjacent similarity suggests a possible transition. In practice, rather than thresholding adjacent similarities directly, we apply kernel change-point detection to the full embedding trajectory. This lets the boundary decision depend on the local structure of the embedding sequence rather than on isolated pairwise differences.

The change-point detector returns boundaries
\[
\mathcal{B} = \{b_1,\dots,b_K\}, \qquad 1 < b_1 < \cdots < b_K < T,
\]
which partition the window sequence into contiguous segments. We use a cosine-kernel change-point objective of the form
\[
\mathcal{L}(\mathcal{B}) =
\sum_{m=0}^{K}
\mathcal{C}\!\left(\mathbf{e}_{b_m+1:b_{m+1}}\right)
+ \beta(T)K,
\]
where $\mathcal{C}(\cdot)$ is the within-segment kernel cost and $\beta(T)$ penalizes excessive segmentation. Throughout, we use $\beta(T)=2\,\mathrm{Var}(s)\log T$, where $s$ is the sequence of cosine similarities between successive window embeddings and $T$ is the total number of windows. This scaling grows slowly with sequence length while aligning the penalty with the similarity spread in each video. We use a standard cosine-kernel CPD implementation \citep{Truong2020CPDImplementation}, with a minimum segment length to avoid degenerate fragments.

The output is a set of adaptive temporal chunks,
\[
\mathcal{S} = \{c_1, c_2, \dots, c_M\},
\]
where each chunk $c_i$ is a contiguous span of windows. This chunking stage is query-independent and task-independent. It is computed once per video and reused throughout the rest of the pipeline.

\subsection{Reusable chunk representation}
\label{subsec:chunk_representation}

Raw change-point outputs are lightly post-processed before being used downstream. Very short chunks are merged into a neighboring chunk when doing so improves local semantic consistency and does not violate a maximum duration constraint. This reduces fragmentation while preserving the content-adaptive structure produced by change-point detection.

Each final chunk is represented by the window embeddings it contains. We do not re-encode a chunk as a new video clip. Instead, we preserve the original window-level embeddings and let each downstream readout decide how to use them. This keeps the video-side computation fixed and avoids discarding fine-grained evidence that may matter for retrieval or frame selection.

The next sections instantiate this shared representation in three settings. Event detection uses chunk edges directly as boundary predictions. Moment retrieval scores the windows inside each chunk against a text query. Long-video VLM reasoning uses the chunks to guide frame selection under a limited visual budget. In all cases, STITCH computes the same query-independent chunk structure once, while only the lightweight readout changes.

%% file: sections/4_experiments.tex
\section{STITCH Variants}
\label{sec:method_variants}

We evaluate a small set of STITCH variants to isolate the main design choices in the pipeline. The default method, \textbf{STITCH}, uses standard kernel change-point detection and encodes each temporal window with four sampled frames. We vary two components: the change-point rule used to form chunks and the number of frames sampled from each window. These variants test whether the default design is necessary and characterize the tradeoff between accuracy and computational cost.

The primary design choice is how boundaries are selected from the embedding stream. \textbf{STITCH} uses standard kernel change-point detection, which evaluates the embedding sequence globally and is less sensitive to isolated local fluctuations than thresholding adjacent similarities directly. \textbf{STITCH-PELT} replaces the standard kernel solver with Pruned Exact Linear Time change-point detection \citep{Killick2011OptimalDO}, giving a linear-time variant of the chunking stage under its pruning conditions. \textbf{STITCH-STD} instead uses a standard-deviation threshold over local embedding changes. This provides a minimal test of whether adjacent similarity drops alone are sufficient, but can be more vulnerable to noisy regions that produce clustered boundaries.

The second design choice is how many frames are sampled from each temporal window before encoding. \textbf{STITCH-2F} uses two frames per window while keeping the rest of the pipeline unchanged, testing whether the method remains effective with a smaller per-window visual budget. Unless otherwise stated, all reported STITCH results use standard kernel change-point detection with four frames per window. Full hyperparameter values are reported in the supplemental (Section~\ref{sec:hyperparams}).

% \red{Figure for this section, maybe we put all plots side by side, a, b, c, d. Bar plots showing performance}

% Section 4: 
% 

% Section 4:
% - Std vs cpd - plot
% - VideoMAE vs internvideo2 bar plot performance (on moment retrieval)
% - Window embedding size (also show performance on moment retrieval)

% TODO: circle back and can add stuff

% Nice to have design choise 1: Semantic Signal generation, Semantic backbone
% \section{Experiments}
% \label{sec:experiments}

% TODO: result tables are currently resized to \textwidth; if the venue style allows, consider wider layouts (e.g.\ \texttt{table*}, modestly larger \verb|\tabcolsep|, or margin tweaks) to reduce cramping.

%% file: sections/5_app1_gebd.tex
% Group all three task-example figures together so they share one float page
% before Figure 5 (the supplementary plots in section 7).

\section{Application 1: Event Detection}
\label{sec:app:event}

For generic event boundary detection, the chunking procedure itself becomes the predictor. After computing the adaptive chunk partition, we treat its boundaries as event boundary predictions. This application therefore requires no task-specific machinery beyond the shared front end. We encode frozen window embeddings, apply cosine-kernel change-point detection, post-process the resulting chunks, and use the final boundaries directly. The central test is whether semantic changes in the pretrained embedding stream align with human-annotated event transitions, which we evaluate on Kinetics-GEBD \citep{shou2021gebd}, a benchmark of generic event boundaries in diverse web videos, and TAPOS \citep{shao2020tapos}, which contains finer procedural sub-action boundaries in Olympic sports videos.

Event detection provides a clean evaluation of the abstraction layer because it removes query scoring and downstream reasoning. If the temporal units are meaningful, their boundaries should coincide with visible changes in action or scene structure. Figure~\ref{fig:event_examples} shows this behavior on two Kinetics-GEBD clips, where predicted boundaries land near interpretable changes in the visible action.

% Notes about fig 2:
% Labels underneath - remove example video
% say videos are from kinetics gebd
% ex video 1 - change frame 1, 
% Maybe just call the Method Stitch

% talk about content detector and put info on it in supplemental materials

%% file: sections/6_app2_mr.tex
\section{Application 2: Moment Retrieval}
\label{sec:app:search}

Video moment retrieval is the task of localizing the temporal span in an untrimmed video that best matches a natural-language query. For this task, STITCH reuses the same chunk structure as a query-independent proposal set. Given a text query $q$, we compute a query embedding
\[
\mathbf{z}_q = f_{\mathrm{text}}(q),
\]
using the same frozen backbone. Each chunk is then scored by comparing the query embedding to the window embeddings inside that chunk. Our default chunk score is the maximum similarity,
\[
\mathrm{score}(c_i, q) = \max_{\mathbf{e}_t \in c_i} \cos(\mathbf{z}_q, \mathbf{e}_t).
\]

This choice reflects the role of a chunk in retrieval. A chunk may contain a relevant sub-event without every window in the chunk being equally aligned to the query. Averaging can wash out this signal, while MaxSim preserves the strongest local match.

Chunk scores are used directly for retrieval. For single-moment settings, we select the highest-scoring chunk or rank chunks by score. When relevant spans extend across neighboring chunks or multiple moments may be present, we optionally group adjacent chunks with similar query scores. The video-side computation remains unchanged. We do not reprocess the full video per query or train a query-conditioned temporal head. The retrieval application only adds query scoring and optional score-based grouping on top of the reusable chunk structure. We evaluate on ActivityNet Captions \citep{krishna2017densecaption}, which pairs untrimmed activity videos with sentence-level temporal annotations, and QVHighlights \citep{Lei2021QVHighlightsDM}, which includes moment-retrieval annotations for shorter, highlight-oriented video segments. Figure~\ref{fig:moment_retrieval_examples} shows examples from ActivityNet Captions.

% TODO: put all the figures all together on one page
% Put temporal abstract layer here
% For the paper with just figures - explain lots of stuff about what we are doing.
% Also have captions refer to other sections - for each of these figures

% Table feedback TOOD:
% Put these tables together - put an x or N/A or unreported 
% We are the only ones that have attempted both
% Huge table 
% In table for each adapation of STITCH put footnotes for each type
% x impossible dash - unreported / unavailable

% Core Design choice 4: Mean vs Max vs GEM. Some results 

%% file: sections/7_app3_vlm.tex
\section{Application 3: Video QA Understanding with VLMs}
\label{sec:app:semantic}

For long-video question answering with VLMs, STITCH is used for frame selection rather than direct prediction. A VLM can only inspect a limited number of frames, so the problem is to allocate that visual budget across the video. Uniform sampling ignores the semantic structure of the video and can waste frames on redundant or irrelevant regions. STITCH instead provides a content-aware chunk structure that can guide which parts of the video are shown to the model. We evaluate this use case on MLVU \citep{Zhou2024MLVUBM}, LongVideoBench \citep{wu2024longvideobench}, and VideoMME \citep{Fu2025VideoMME}, three long-video multiple-choice QA benchmarks where STITCH controls only which visual evidence is supplied to the downstream VLM.

We use two simple chunk-based selection rules. The first, \textbf{Greedy}, uses the STITCH chunks directly. Given a question, we embed the question with the frozen text encoder and score each chunk by the maximum similarity between the query embedding and the window embeddings inside that chunk. We then select chunks in decreasing score order and pass one representative frame from each selected chunk to the VLM. This is the most direct use of STITCH for VLM reasoning: the same chunks computed by the shared temporal abstraction layer become the units for allocating frames.

The second rule uses a chunk-aware variant of Maximal Marginal Relevance (\textbf{MMR}) \citep{carbonell1998mmr}. MMR was originally introduced to reduce redundancy and increase diversity when reranking retrieved documents. We adapt the same principle to frame selection, where the goal is to choose frames that are relevant to the question without wasting budget on repeated visual evidence. Unlike Greedy, this rule scores and selects individual windows rather than whole chunks. Each selected window contributes one representative frame to the VLM. This allows the selector to choose nearby frames when they are useful, while still discouraging repeated selections from the same semantic chunk.

Let $\mathcal{A}$ denote the set of selected window indices, and let $g(i)$ denote the chunk containing window $i$. Each candidate window is scored as
\[
\mathrm{MMR}_{\mathrm{chunk}}(i) =
\lambda \, \cos(\mathbf{e}_i, \mathbf{z}_q)
-
(1-\lambda)
\Big[
\max_{j \in \mathcal{A}} \cos(\mathbf{e}_i, \mathbf{e}_j)
+
\gamma \cdot [g(i) \in \mathcal{R}_{\mathcal{A}}]
\Big],
\]
where $\mathcal{R}_{\mathcal{A}} = \{g(j) : j \in \mathcal{A}\}$ is the set of chunks already represented. The first term favors windows relevant to the question. The second term penalizes redundancy with already selected windows, and the chunk penalty discourages spending multiple frames on the same semantic segment. The parameters $\lambda$ and $\gamma$ control the relevance--diversity tradeoff and the strength of the same-chunk penalty. We use one fixed setting of these parameters for all long-video VLM benchmarks, rather than tuning them separately for each model--dataset pair. This setting is selected once and then applied unchanged to MLVU, LongVideoBench, and VideoMME, as specified in Appendix Sec.~\ref{sec:hyperparams}. Frames are selected greedily under this score until the budget is filled.

Both rules use the same frozen embeddings and the same STITCH chunk structure. They differ only in how strictly the frame budget is tied to chunk boundaries. We compare these strategies in the supplemental results (Sec.~\ref{sec:supp_full_results}) and report the MMR variant in the main table because it tends to perform slightly better. The use of one shared hyperparameter setting across all VLM QA benchmarks supports the interpretation of STITCH as a reusable selection front end rather than a benchmark-specific selection heuristic. Figure~\ref{fig:vlm_qa_examples} shows an illustrative QA example, including the chunk timeline and the frames supplied to the VLM under a fixed budget.

% TODO: put a box around the quw

% Table~\ref{tab:vlm_qa} reports \textbf{Base} (uniform), \textbf{+Method} (with frame selection), and \textbf{$\Delta$} (gain) on MLVU dev (M-Avg, \citep{Zhou2024MLVUBM}), LongVideoBench val, and VideoMME test.

% TODO: MLVU---add mean (or representative) video duration in this section or in the table footnote/caption once computed from the benchmark.
% TODO: VLM QA table---resize/wrap rules for final camera-ready layout; optional row highlighting / $\Delta$ coloring to match comparison tables.

% Figure~\ref{fig:vlm_qa_plots} shows supplementary diagnostics for Qwen3-VL-8B.

%% file: full_page_methods_fig.tex
\begin{figure}[p]
    \centering
    \includegraphics[width=\linewidth]{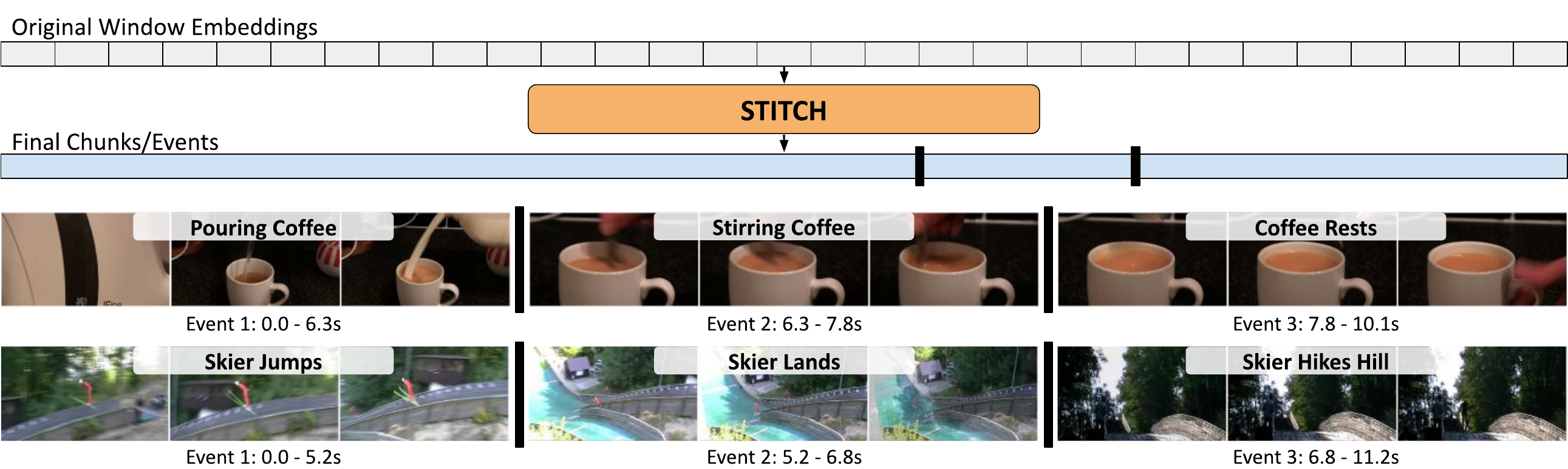}
    \caption{\textbf{\textit{GEBD}: STITCH boundaries reveal interpretable event transitions.} In the event detection setting of Section~\ref{sec:app:event}, STITCH uses chunk boundaries from the shared temporal abstraction as event boundary predictions. The text labels are manually added descriptions of the visible events, not outputs of the model; they are included to clarify what the predicted cuts separate in the video.}
    \label{fig:event_examples}
    
    \vspace{2em}
    \includegraphics[width=\linewidth]{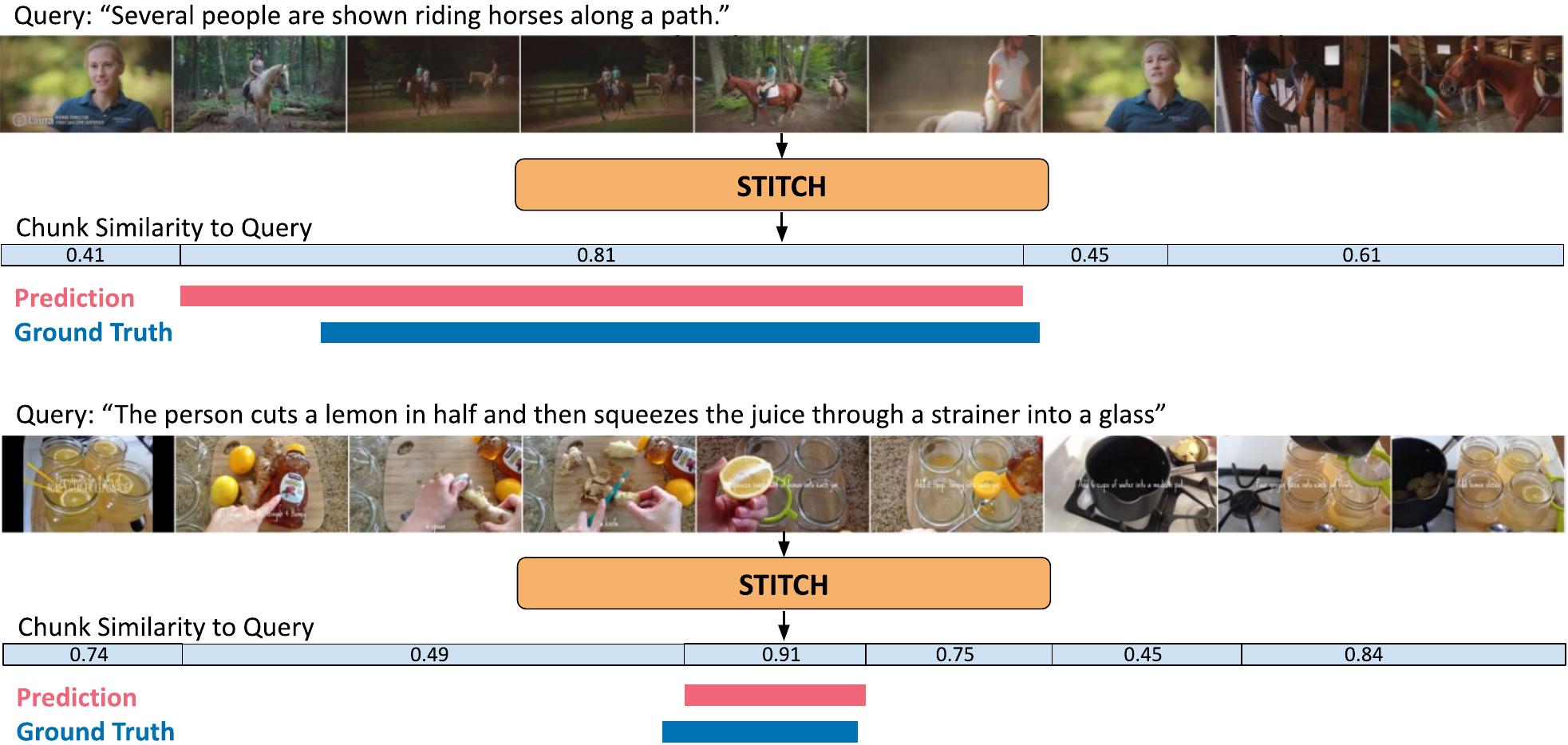}
    \caption{\textbf{\textit{Moment Retrieval}: Language queries ground reusable chunks into relevant video spans.} For the retrieval readout in Section~\ref{sec:app:search}, STITCH chunks the video once and scores the resulting spans against a language query in the shared embedding space. This turns the task-independent chunk timeline into query-specific interval predictions while keeping the video encoding fixed.}
    \label{fig:moment_retrieval_examples}
    \vspace{2em}

    \includegraphics[width=\linewidth]{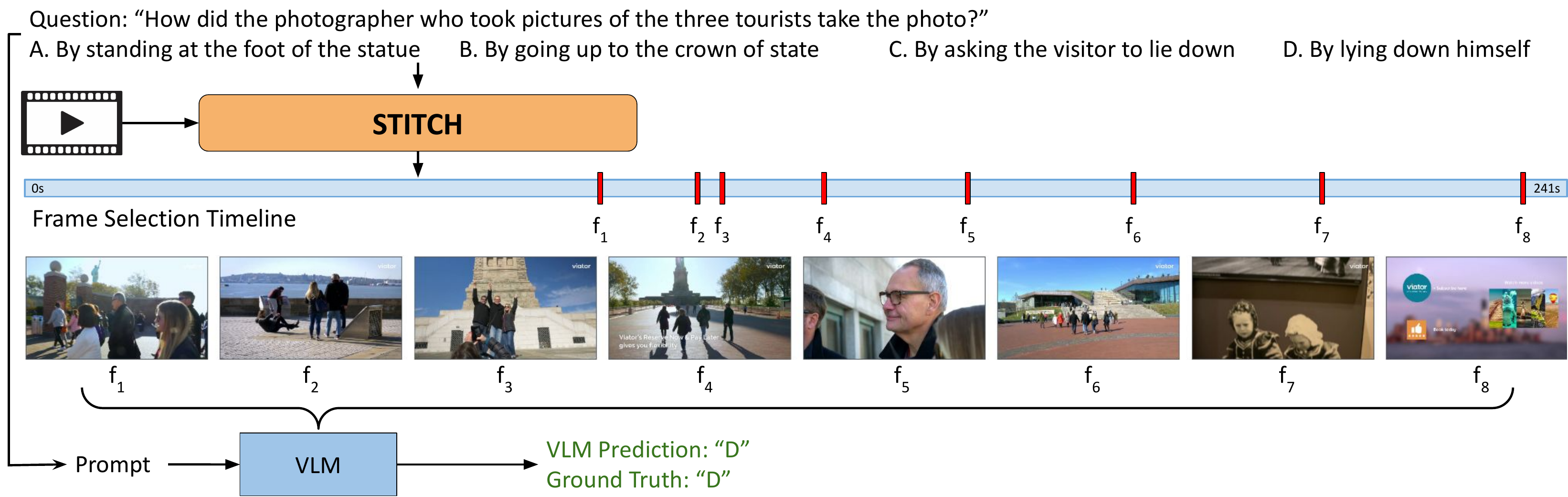}
    \caption{\textbf{\textit{VLM Frame Selection}: Semantic chunks select compact evidence for better VLM reasoning.} In Section~\ref{sec:app:semantic}, STITCH first partitions the video into chunks, scores those chunks by their relevance to the query, and then selects a small set of frames from those chunks under a fixed visual budget. The selected frames cover distinct parts of the video while remaining relevant to the question, giving the VLM targeted evidence for video reasoning.}
    \label{fig:vlm_qa_examples}
\end{figure}

%% file: main_results_table.tex
\begin{table}[t]
\centering
\small
\setlength{\tabcolsep}{4.0pt}
\renewcommand{\arraystretch}{1.10}
\caption{\textbf{Cross-task summary of specialized baselines and STITCH.}
The table groups prior methods by their reported task scope, so the \textsc{N/A} blocks indicate tasks outside a method's evaluated setting rather than missing results. STITCH is evaluated across all three task families using the same training-free temporal abstraction front end. For event detection and moment retrieval, we report the representative metrics shown in the column headers; for long-video VLM QA, $\Delta$ denotes the accuracy gain over uniform frame sampling. All values are percentages. VLM results use LLaVA-OV 7B~\citep{li2025llavaonevision} with eight input frames and the MMR-based STITCH selector. Full metrics, uniform baselines for the VLM deltas, and additional STITCH variants are reported in Section~\ref{sec:supp_full_results}. Unified localization frameworks are discussed in Section~\ref{sec:prior} rather than reported as single rows because their benchmark suites only partially overlap with ours. More importantly, they unify localization-style prediction tasks, whereas STITCH also includes retrieval and frame-budgeted VLM reasoning.
\vspace{0.35em}\\
\emph{Types:} Sup.~= supervised, Unsup.~= unsupervised, ZS~= zero-shot transfer, TF~= training-free.
\vspace{0.5em}}
\label{tab:summary_all_tasks}
\resizebox{\textwidth}{!}{%
\begin{tabular}{c l c | cc | cc | ccc}
\hline
& & &
\multicolumn{2}{c|}{Generic event boundary detection} &
\multicolumn{2}{c|}{Moment retrieval} &
\multicolumn{3}{c}{Long-video VLM QA} \\
\cline{4-10}
Focus & Method & Type &
Kinetics-GEBD & TAPOS~\citep{shao2020tapos} &
ActivityNet & QVHighlights &
MLVU~\citep{Zhou2024MLVUBM} & LongVideoBench~\citep{wu2024longvideobench} & VideoMME~\citep{Fu2025VideoMME} \\
& & &
Avg F1 & Avg F1 &
R@1@.5 & R@1@.5 &
$\Delta$ & $\Delta$ & $\Delta$ \\
\hline

\multirow{7}{*}{\rotatebox[origin=c]{90}{\small Event detection}}
    & BMN~\citep{lin2019bmn, shou2021gebd}
    & ZS
    & 22.3 & 15.4
    & \multicolumn{2}{c|}{\multirow{7}{*}{\normalsize\textsc{N/A}}}
    & \multicolumn{3}{c}{\multirow{7}{*}{\normalsize\textsc{N/A}}} \\
& TCN~\citep{lea2016segmental, shou2021gebd}
    & Sup.
    & 68.5 & 33.0
    & \multicolumn{2}{c|}{}
    & \multicolumn{3}{c}{} \\
& PC~\citep{shou2021gebd}
    & Sup.
    & 81.7 & 64.3
    & \multicolumn{2}{c|}{}
    & \multicolumn{3}{c}{} \\
& Temporal Perceiver~\citep{tan2023temporalperceiver}
    & Sup.
    & 86.0 & 73.2
    & \multicolumn{2}{c|}{}
    & \multicolumn{3}{c}{} \\
& DyBDet~\citep{zheng2024dybdet}
    & Sup.
    & \textbf{89.0} & 74.7
    & \multicolumn{2}{c|}{}
    & \multicolumn{3}{c}{} \\
& DiffGEBD~\citep{hwang2025diffgebd}
    & Sup.
    & 87.5 & \textbf{75.2}
    & \multicolumn{2}{c|}{}
    & \multicolumn{3}{c}{} \\
& FlowGEBD~\citep{venkatesh2024flowgebd}
    & Unsup.
    & 84.5 & 62.3
    & \multicolumn{2}{c|}{}
    & \multicolumn{3}{c}{} \\

\noalign{\vskip 2pt}
\hline
\noalign{\vskip 2pt}

\multirow{8}{*}{\rotatebox[origin=c]{90}{\small Moment retrieval}}
    & SCDM~\citep{yuan2019scdm}
    & Sup.
    & \multicolumn{2}{c|}{\multirow{8}{*}{\normalsize\textsc{N/A}}}
    & 36.8 & \mbox{---}
    & \multicolumn{3}{c}{\multirow{8}{*}{\normalsize\textsc{N/A}}} \\
& 2D-TAN~\citep{luo2020learning2dtemporal}
    & Sup.
    & \multicolumn{2}{c|}{}
    & 44.5 & \mbox{---}
    & \multicolumn{3}{c}{} \\
& SimBase~\citep{bao2024simbase}
    & Sup.
    & \multicolumn{2}{c|}{}
    & \textbf{49.4} & \mbox{---}
    & \multicolumn{3}{c}{} \\
& TFVTG~\citep{zheng2024tfvtg}
    & TF
    & \multicolumn{2}{c|}{}
    & 27.0 & \mbox{---}
    & \multicolumn{3}{c}{} \\
& Moment-DETR~\citep{Lei2021QVHighlightsDM}
    & Sup.
    & \multicolumn{2}{c|}{}
    & \mbox{---} & 53.9
    & \multicolumn{3}{c}{} \\
& QD-DETR~\citep{Moon_2023_CVPR}
    & Sup.
    & \multicolumn{2}{c|}{}
    & \mbox{---} & 62.7
    & \multicolumn{3}{c}{} \\
& TR-DETR~\citep{Sun_2024_AAAI_TRDETR}
    & Sup.
    & \multicolumn{2}{c|}{}
    & \mbox{---} & 67.1
    & \multicolumn{3}{c}{} \\
& SG-DETR~\citep{hou2024sgdetr}
    & Sup.
    & \multicolumn{2}{c|}{}
    & \mbox{---} & \textbf{72.8}
    & \multicolumn{3}{c}{} \\

\noalign{\vskip 2pt}
\hline
\noalign{\vskip 2pt}

\multirow{6}{*}{\rotatebox[origin=c]{90}{\small VLM QA}}
    & Frame-Voyager~\citep{yu2025framevoyager}
    & Sup.
    & \multicolumn{2}{c|}{\multirow{6}{*}{\normalsize\textsc{N/A}}}
    & \multicolumn{2}{c|}{\multirow{6}{*}{\normalsize\textsc{N/A}}}
    & +7.1 & \mbox{---} & +4.2 \\
& KFC~\citep{fang2026threading}
    & TF
    & \multicolumn{2}{c|}{}
    & \multicolumn{2}{c|}{}
    & +7.7 & +1.1 & +2.1 \\
& BOLT~\citep{liu2025bolt}
    & TF
    & \multicolumn{2}{c|}{}
    & \multicolumn{2}{c|}{}
    & +4.5 & +1.4 & +2.3 \\
& AKS~\citep{tang2025adaptive}
    & TF
    & \multicolumn{2}{c|}{}
    & \multicolumn{2}{c|}{}
    & +4.3 & +4.2 & +4.1 \\
& Frame-Oracle~\citep{li2026frameoracle}
    & Sup.
    & \multicolumn{2}{c|}{}
    & \multicolumn{2}{c|}{}
    & +4.5 & +1.7 & +3.7 \\
& WFS-SB~\citep{chen2026wfsb}
    & TF
    & \multicolumn{2}{c|}{}
    & \multicolumn{2}{c|}{}
    & \textbf{+8.6} & \textbf{+5.6} & \textbf{+5.2} \\

\noalign{\vskip 2pt}
\hline
\noalign{\vskip 2pt}

\multirow{4}{*}{\rotatebox[origin=c]{90}{\small All}}
    & \textbf{STITCH}
    & TF
    & 83.9 & 44.8
    & 32.9 & 64.6
    & +8.5 & +2.8 & +3.0 \\
& \textbf{STITCH-PELT}
    & TF
    & 80.1 & 28.8
    & 31.7 & 55.8
    & +7.7 & +1.9 & +3.0 \\
& \textbf{STITCH-STD}
    & TF
    & 47.2 & 36.2
    & 27.2 & 35.0
    & +7.1 & +2.8 & +3.2 \\
& \textbf{STITCH-2F}
    & TF
    & 83.5 & 44.2
    & 32.0 & 63.4
    & +8.2 & +2.4 & +2.4 \\
\hline
\end{tabular}%
}
\vspace{-0.0em}
\end{table}

%% file: sections/8_discussion.tex
\section{Discussion}
\label{sec:discussion}
Table~\ref{tab:summary_all_tasks} shows that STITCH is strongest when compared with training-free, zero-shot, and unsupervised methods, while remaining competitive with many specialized systems. This distinction is important. Supervised methods can learn the annotation conventions and temporal granularity of a target benchmark; STITCH instead extracts one chunk structure from a frozen video-text embedding stream and reuses it across tasks. This reflects a stricter notion of generality, where a method is general not because it can be retrained for many benchmarks, but because the same video representation can be reused without task-specific adaptation. The benchmark set stresses this reuse across visible event changes in Kinetics-GEBD, finer sub-action boundaries in TAPOS, sentence-level localization in ActivityNet Captions, highlight-oriented retrieval in QVHighlights, and frame-budget allocation in long-video QA.

\tightparagraph{Event detection.}
Event detection is the most direct test of the chunking step. On Kinetics-GEBD, STITCH reaches 83.9 Avg F1, close to supervised boundary detectors and competitive with unsupervised methods despite using no boundary labels. TAPOS exposes a finer-grained regime where benchmark-specific supervision is more valuable. STITCH reaches 44.8 Avg F1, well above the zero-shot BMN baseline but below supervised methods trained for procedural sub-action boundaries. This gap clarifies the tradeoff: reusable chunks capture broad semantic transitions well, while fine sub-action boundaries still benefit from task-specific learning.

\tightparagraph{Moment retrieval.}
Moment retrieval tests whether query-independent chunks can act as temporal proposals. On ActivityNet Captions, it outperforms TFVTG, the training-free baseline in Table~\ref{tab:summary_all_tasks}. On QVHighlights, STITCH reaches 64.6 R@1@0.5, exceeding Moment-DETR and QD-DETR despite using no moment annotations or query-conditioned temporal head. This is a useful signal because QVHighlights is a supervised moment-retrieval benchmark. STITCH is not merely competitive with other training-free methods, but recovers enough proposal structure from frozen video-text embeddings to surpass some learned retrieval systems on their own evaluation setting. Stronger supervised systems still perform better, suggesting that task-specific learning improves proposal ranking and boundary refinement, but the gap is smaller than one might expect for a reusable query-independent front end.

\begin{figure}[t]
    \centering
    \begin{subfigure}[t]{0.33\textwidth}
        \centering
        \includegraphics[width=\linewidth]{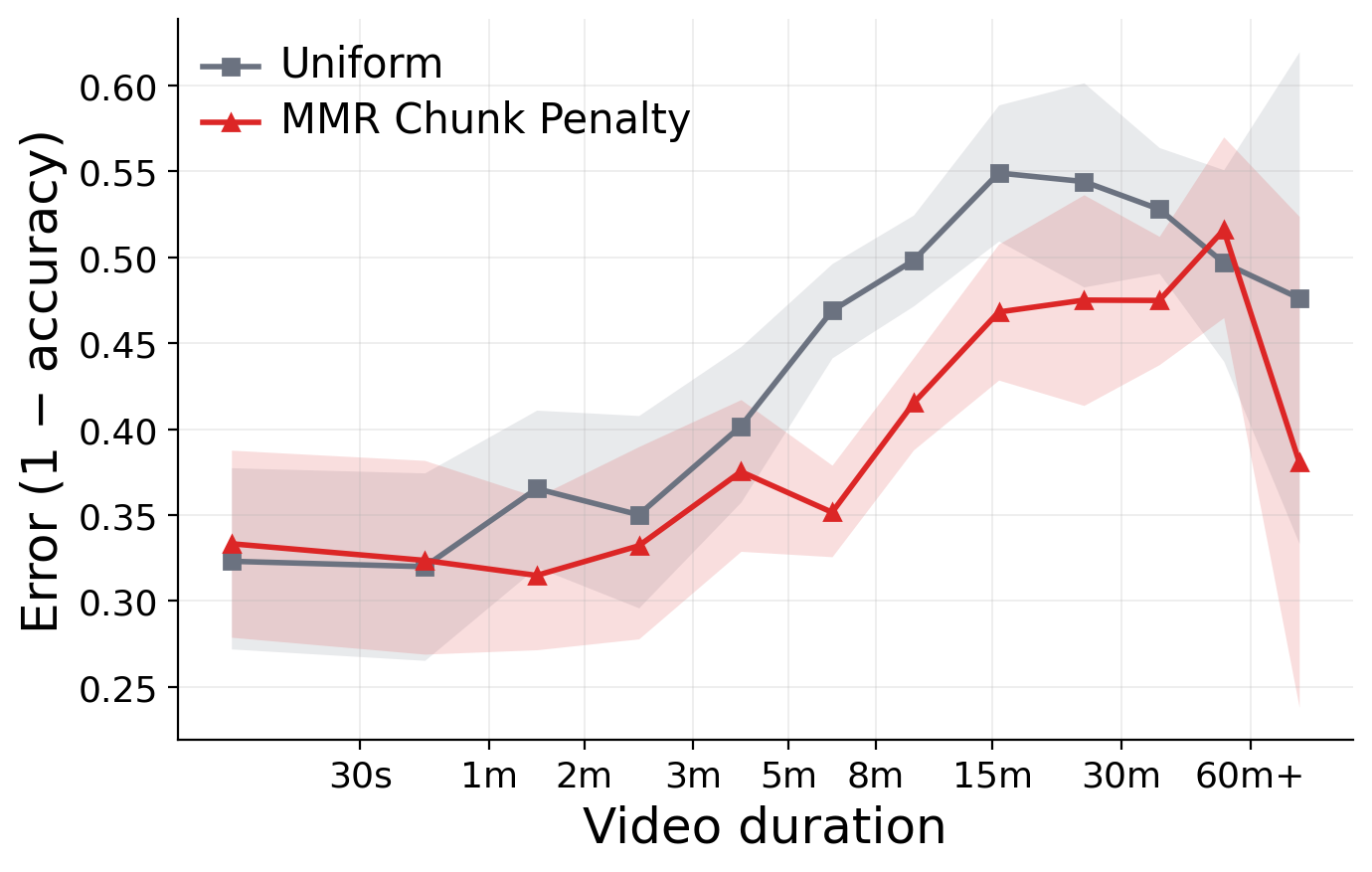}
        \label{fig:vlm_qa_plot_agg}
    \end{subfigure}%
    \hfill
    \begin{subfigure}[t]{0.33\textwidth}
        \centering
        \includegraphics[width=\linewidth]{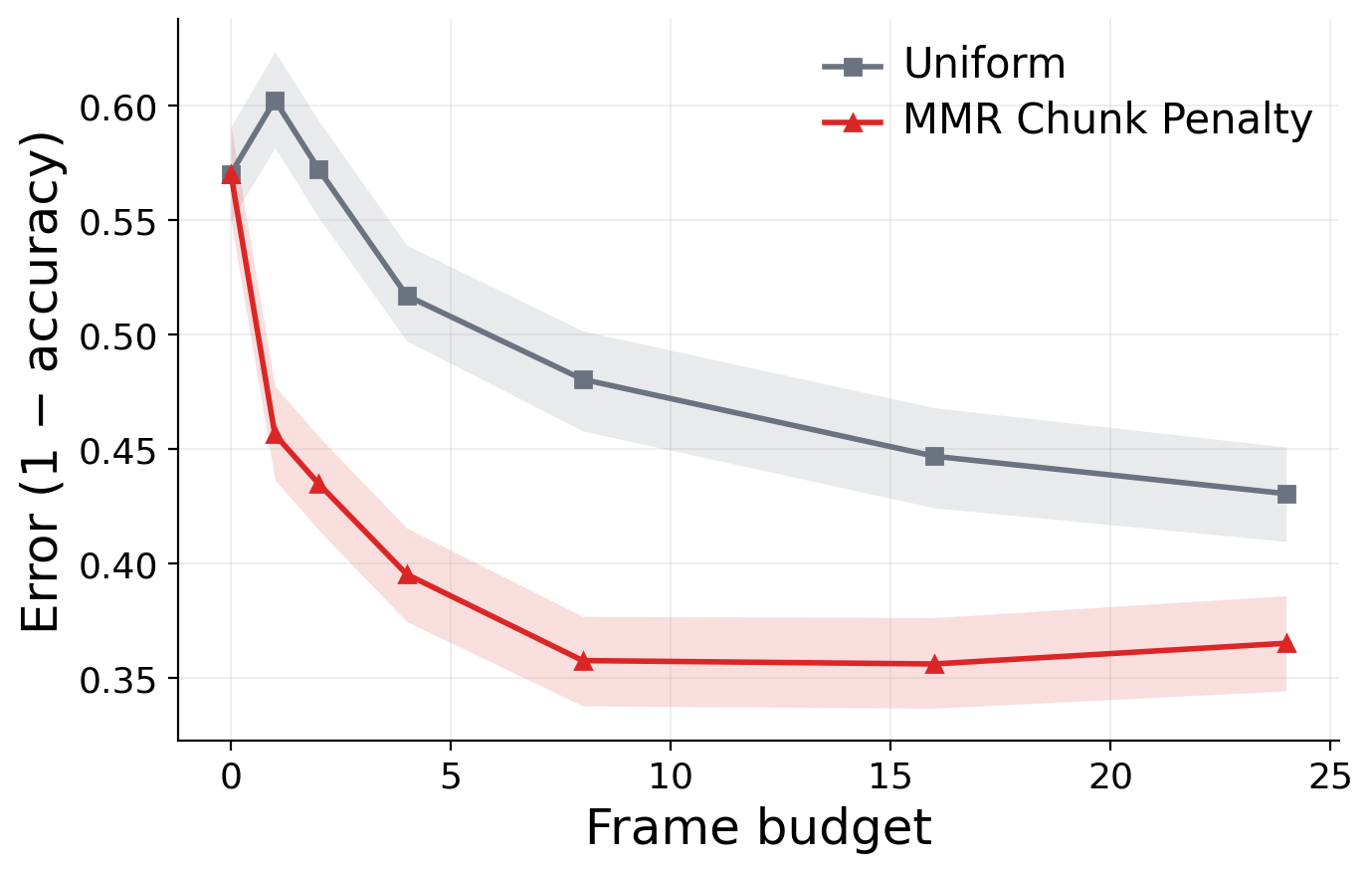}
        \label{fig:vlm_qa_plot_mlvu_overall}
    \end{subfigure}%
    \hfill
    \begin{subfigure}[t]{0.33\textwidth}
        \centering
        \includegraphics[width=\linewidth]{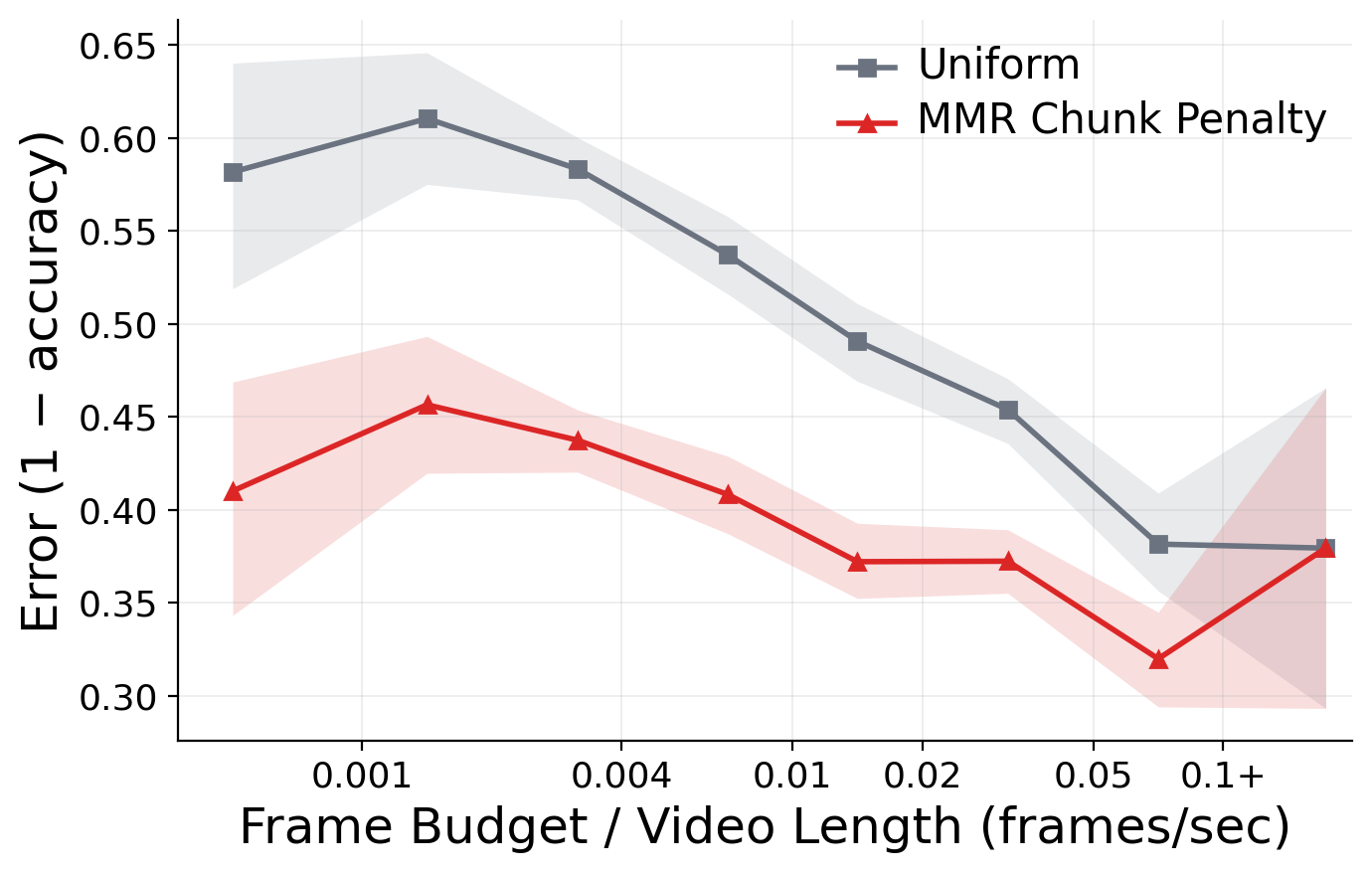}
        \label{fig:vlm_qa_plot_mlvu_duration}
    \end{subfigure}
    % \caption{\textbf{STITCH gains are largest when frame density is lowest.}
    % Complementing the VLM QA readout in Section~\ref{sec:app:semantic}, these diagnostics compare STITCH-based selection with uniform sampling for Qwen3-VL-8B~\citep{Qwen3-VL}. \textbf{Left}: error across video durations, aggregated over VideoMME, MLVU, and LongVideoBench for an 8-frame budget. \textbf{Center}: error on MLVU as a function of frame budget. \textbf{Right}: error on MLVU as a function of normalized frame budget in frames/sec.}
    \caption{\textbf{STITCH gains are largest when frame density is lowest.}
    These diagnostics compare STITCH-based selection with uniform sampling for Qwen3-VL-8B~\citep{Qwen3-VL}. \textbf{Left}: error by video duration, aggregated over VideoMME, MLVU, and LongVideoBench at an 8-frame budget. \textbf{Center}: MLVU error by frame budget. \textbf{Right}: MLVU error by normalized frame budget in frames/sec.}
    \label{fig:vlm_qa_plots}
    \vspace{-0.2em}
\end{figure}

\tightparagraph{Long-video QA.}
Long-video QA tests whether reusable chunks help allocate a small visual budget of frames the VLM can inspect.  With eight input frames, the chunk-aware MMR selector improves over uniform sampling on MLVU, LongVideoBench, and VideoMME, using the same hyperparameters across all three benchmarks. Figure~\ref{fig:vlm_qa_plots} provides additional analysis showing that STITCH improves frame selection across longer videos and varied frame-budget settings. The gains are largest on MLVU and remain positive on the longer benchmarks, suggesting that semantic chunks provide a useful prior for selecting sparse visual evidence. STITCH is close to the strongest method on MLVU in Table~\ref{tab:summary_all_tasks}, while trailing more specialized frame-selection methods on LongVideoBench and VideoMME.

\tightparagraph{Method variants and limitations.}
The variant results suggest that the gains are not tied to one narrow configuration. STITCH-2F remains close to the four-frame default, including 83.5 versus 83.9 on Kinetics-GEBD and 63.4 versus 64.6 on QVHighlights, reducing per-window visual cost while preserving most of the benefit. The change-point solver matters more. PELT remains strong in several settings, but standard kernel change-point detection gives the most consistent performance, while STD thresholding shows that local similarity drops alone are not enough for reliable temporal units. STITCH remains limited by the frozen embedding space: subtle transitions can be missed, nuisance motion can cause over-segmentation, and a single-scale chunk structure cannot capture every level of temporal organization. These limitations point toward hierarchical chunking, stronger backbones, and streaming variants that maintain an online memory of semantic chunks.

% TODO: exapnd on Long Video QA

% \subsection{Conclusion}

% We introduce STITCH, a training-free method for semantic temporal chunking with a frozen video-text backbone. The same chunk structure supports event detection, moment retrieval, and long-video QA frame selection without task-specific training. Across these settings, STITCH remains competitive with more specialized methods while keeping the video-side computation reusable.

% These results support temporal chunking as a practical abstraction for general video understanding. A video does not always need to be processed as a dense sequence of frames or as fixed windows. It can first be converted into semantic units, then searched, scored, sampled, or stored according to the downstream task. As video models become stronger and long-context applications become more common, this kind of reusable temporal structure may become an increasingly useful interface between raw video and higher-level reasoning systems.

%% file: supplemental.tex
\section{Full Benchmark Results}
\label{sec:supp_full_results}

\subsection{GEBD}
% TODO: report exact number of Kinetics-GEBD val videos evaluated; count varies with YouTube availability at download time
\begin{table}[ht]
\centering
\caption{Generic event boundary detection on Kinetics-GEBD (val) and TAPOS (val). F1 scores (\%) at selected relative-distance thresholds and averaged over all ten (0.05--0.50). Sup.~= supervised, Unsup.~= unsupervised, ZS~= zero-shot transfer (trained on a different dataset and applied without retraining), TF~= training-free.}
\label{tab:gebd}
\resizebox{\textwidth}{!}{%
\begin{tabular}{ll | ccccc | ccccc}
\toprule
& & \multicolumn{5}{c|}{Kinetics-GEBD} & \multicolumn{5}{c}{TAPOS} \\
\cmidrule(lr){3-7} \cmidrule(lr){8-12}
Method & Type & F1@.05 & F1@.10 & F1@.20 & F1@.30 & Avg & F1@.05 & F1@.10 & F1@.20 & F1@.30 & Avg \\
\midrule
BMN~\citep{lin2019bmn, shou2021gebd}          & ZS     & 18.6 & 20.4 & 22.0 & 23.0 & 22.3 & 12.8 & 14.1 & 15.2 & 15.9 & 15.4 \\
TCN~\citep{lea2016segmental, shou2021gebd}    & Sup.   & 58.8 & 65.7 & 69.1 & 70.3 & 68.5 & 23.7 & 31.2 & 33.9 & 34.4 & 33.0 \\
PC~\citep{shou2021gebd}                      & Sup.   & 62.5 & 75.8 & 82.9 & 85.3 & 81.7 & 52.2 & 59.5 & 64.7 & 66.6 & 64.3 \\
Temporal Perceiver~\citep{tan2023temporalperceiver} & Sup.   & 74.8 & 82.8 & 86.6 & 87.9 & 86.0 & 55.2 & 66.3 & 73.8 & 76.5 & 73.2 \\
DyBDet~\citep{zheng2024dybdet}               & Sup.   & \textbf{79.6} & \textbf{85.8} & \textbf{89.3} & \textbf{90.7} & \textbf{89.0} & 62.5 & 70.1 & 75.6 & \textbf{77.2} & 74.7 \\
DiffGEBD~\citep{hwang2025diffgebd}           & Sup.   & 78.4 & 84.8 & 87.9 & 89.1 & 87.5 & \textbf{65.8} & \textbf{71.8} & \textbf{75.7} & 77.0 & \textbf{75.2} \\
FlowGEBD~\citep{venkatesh2024flowgebd}       & Unsup. & 71.3 & 82.8 & 85.8 & 86.4 & 84.5 & 37.5 & 50.2 & 62.4 & 67.7 & 62.3 \\
\textbf{STITCH (Ours) }                               & TF     & 69.90 & 79.42 & 84.56 & 86.55 & 83.89 & 31.2 & 39.6 & 45.1 & 47.2 & 44.8 \\
\bottomrule
\end{tabular}%
}
\end{table}

\subsection{Moment Retrieval}

Tables~\ref{tab:moment_retrieval_anet} and~\ref{tab:moment_retrieval_qvh} show the full moment retrieval results alongside four STITCH variants corresponding to different chunk scoring rules. Beyond max-similarity, we evaluate mean pooling; generalized mean (GEM)~\citep{RadenovicTC19}, which applies a coordinate-wise power-$p$ mean and interpolates between the arithmetic mean ($p{=}1$) and the coordinate-wise maximum ($p{\to}\infty$); and coherence-weighted mean, which weights each window by $\sum_j \exp(\tau\,e_i^\top e_j)$ so that windows more similar to the rest of the chunk receive higher weight, acting as a soft medoid. All four variants perform comparably, suggesting the retrieval pipeline is not sensitive to the specific pooling rule.

\begin{table}[ht]
\centering
\caption{Language-based moment retrieval on ActivityNet Captions (val split, 17,505 queries). All values are percentages. R@1 IoU=$\tau$ denotes recall at rank 1 with IoU threshold $\tau$. $^\dagger$GEM with $p{=}3$; $^\ddagger$coherence-weighted mean with $\tau{=}5$. Sup.~= supervised, TF~= training-free.}
\label{tab:moment_retrieval_anet}
\begin{tabular}{ll | cccc}
\toprule
Method & Type & R@1 IoU=.3 & R@1 IoU=.5 & R@1 IoU=.7 & mIoU \\
\midrule
SCDM~\citep{yuan2019scdm}                   & Sup. & 54.80 & 36.75 & 19.86 & --    \\
2D-TAN~\citep{luo2020learning2dtemporal}    & Sup. & 59.45 & 44.51 & 26.54 & --    \\
SimBase~\citep{bao2024simbase}              & Sup. & \textbf{63.98} & \textbf{49.35} & \textbf{30.48} & \textbf{47.07} \\
\midrule
TFVTG~\citep{zheng2024tfvtg}                & TF   & 49.34 & 27.02 & 13.39 & 34.10 \\
STITCH-Max (Ours)                           & TF   & 52.25 & 32.85 & 17.87 & 37.40 \\
STITCH-Mean (Ours)                          & TF   & 51.54 & 32.51 & 17.95 & 37.10 \\
STITCH-GEM$^\dagger$ (Ours)                 & TF   & 51.30 & 32.33 & 17.95 & 36.91 \\
STITCH-Coherence$^\ddagger$ (Ours)          & TF   & 51.57 & 32.56 & 18.01 & 37.14 \\
\bottomrule
\end{tabular}
\end{table}

\begin{table}[ht]
\centering
\caption{Language-based moment retrieval on QVHighlights (val split). All values are percentages. $^\dagger$GEM with $p{=}12$; $^\ddagger$coherence-weighted mean with $\tau{=}5$. Sup.~= supervised, TF~= training-free.}
\label{tab:moment_retrieval_qvh}
\begin{tabular}{ll | ccc}
\toprule
Method & Type & R1@.5 & R1@.7 & mAP avg \\
\midrule
Moment-DETR~\citep{Lei2021QVHighlightsDM} & Sup.           & 53.94 & 34.84 & 32.20 \\
QD-DETR~\citep{Moon_2023_CVPR}            & Sup.           & 62.68 & 46.66 & 41.22 \\
TR-DETR~\citep{Sun_2024_AAAI_TRDETR}      & Sup.           & 67.10 & 51.48 & 45.09 \\
SG-DETR~\citep{hou2024sgdetr}              & Sup.           & \textbf{72.80} & \textbf{59.50} & \textbf{55.60} \\
\midrule
STITCH-Max (Ours)                          & TF             & 64.58 & 49.10 & 41.76 \\
STITCH-Mean (Ours)                         & TF             & 64.97 & 49.94 & 43.11 \\
STITCH-GEM$^\dagger$ (Ours)                & TF             & 65.29 & 49.81 & 43.23 \\
STITCH-Coherence$^\ddagger$ (Ours)         & TF             & 64.97 & 50.00 & 43.20 \\
\bottomrule
\end{tabular}
\end{table}

\subsection{VLM QA}

VLM inference and prompt formatting follow the evaluation setup of WFS-SB~\citep{chen2026wfsb}, which is itself built on the lmms-eval framework~\citep{zhang2024lmmsevalrealitycheckevaluation, lmms_eval2024}.

\begin{table}[ht]
\centering
\small
\setlength{\tabcolsep}{3.5pt}
\caption{Multiple-choice VLM accuracy with eight input frames (\%). \textbf{Base}: uniform; \textbf{+Method}: with selection; \textbf{$\Delta$}: gain. Splits: MLVU dev (M-Avg) \citep{Zhou2024MLVUBM}, LongVideoBench val \citep{wu2024longvideobench}, VideoMME test \citep{Fu2025VideoMME}. VLMs evaluated: LLaVA-OV \citep{li2025llavaonevision} and Qwen3-VL \citep{Qwen3-VL}. $^{\dagger}$~As reported in the method's original paper; other LLaVA-OV baselines re-evaluated in that setup. $^{\ddagger}$~Training-based. Dashes: LongVideoBench missing for Frame-Voyager in that table.}
\label{tab:vlm_qa}
\resizebox{\textwidth}{!}{%
\begin{tabular}{@{}>{\raggedright\arraybackslash}p{2.0cm} l c c | c c c | c c c | c c c@{}}
\toprule
Model & Method & Size & Fr. & \multicolumn{3}{c|}{MLVU} & \multicolumn{3}{c|}{LongVideoBench} & \multicolumn{3}{c@{}}{VideoMME} \\
\cmidrule(lr){5-7} \cmidrule(lr){8-10} \cmidrule(lr){11-13}
 & & & & Base & +Method & $\Delta$ & Base & +Method & $\Delta$ & Base & +Method & $\Delta$ \\
\midrule
\multirow{2}{*}{\parbox[c]{1.9cm}{\raggedright Qwen3-VL-8B}} & \textbf{greedy (ours)} & 8B & 8 & 50.89 & 60.59 & +9.70 & 52.43 & 54.67 & +2.24 & 55.63 & 57.70 & +2.07 \\
& \textbf{MMR penalty (ours)} & 8B & 8 & 50.89 & 62.43 & +11.54 & 52.43 & 55.80 & +3.37 & 55.63 & 58.59 & +2.96 \\
\midrule
\multirow{8}{*}{\parbox[c]{1.9cm}{\raggedright LLaVA-OV}} & Frame-Voyager$^{\dagger\ddagger}$~\citep{yu2025framevoyager} & 7B & 8 & 58.5 & 65.6 & +7.1 & \mbox{---} & \mbox{---} & \mbox{---} & 53.3 & 57.5 & +4.2 \\
& KFC$^{\dagger}$~\citep{fang2026threading} & 7B & 8 & 58.5 & 66.2 & +7.7 & 54.5 & 55.6 & +1.1 & 53.3 & 55.4 & +2.1 \\
& BOLT$^{\dagger}$~\citep{liu2025bolt} & 7B & 8 & 58.9 & 63.4 & +4.5 & 54.2 & 55.6 & +1.4 & 53.8 & 56.1 & +2.3 \\
& AKS~\citep{tang2025adaptive} & 7B & 8 & 58.6 & 62.9 & +4.3 & 54.2 & 58.4 & +4.2 & 54.1 & 58.2 & +4.1 \\
& Frame-Oracle$^{\dagger\ddagger}$~\citep{li2026frameoracle} & 7B & 8 & 58.4 & 62.9 & +4.5 & 54.3 & 56.0 & +1.7 & 53.8 & 57.5 & +3.7 \\
& WFS-SB~\citep{chen2026wfsb} & 7B & 8 & 58.6 & 67.2 & +8.6 & 54.2 & 59.8 & +5.6 & 54.1 & 59.3 & +5.2 \\
\cmidrule(lr){2-13}
& \textbf{greedy (ours)} & 7B & 8 & 58.28 & 63.30 & +5.02 & 54.75 & 55.72 & +0.97 & 53.15 & 55.44 & +2.29 \\
& \textbf{MMR penalty (ours)} & 7B & 8 & 58.28 & 66.77 & +8.49 & 54.75 & 57.59 & +2.84 & 53.15 & 56.17 & +3.02 \\
\bottomrule
\end{tabular}%
}
\end{table}
% TODO: change to 1 decimal point
% TODO: put LLAVA over qwen

\section{Hyperparameters}
\label{sec:hyperparams}

\paragraph{Chunking and post-processing.}
The window sampling interval (Section~\ref{subsec:dense_embeddings}) is $0.5\,$s for GEBD and $2.0\,$s for moment retrieval and VLM QA; each window spans exactly one sampling interval. The minimum segment length passed to the change-point solver is 2 windows, corresponding to $1.0\,$s for GEBD and $4.0\,$s for the other tasks.

The change-point penalty $\beta(T)$ is as defined in Section~\ref{subsec:chunking}. For moment retrieval and VLM QA we use the automatic formula without modification. For event boundary detection, where the number of predicted boundaries directly affects recall--precision balance, we fix $\beta$ to a constant tuned on the training split of each dataset: $\beta = 0.03$ on Kinetics-GEBD and $\beta = 0.07$ on TAPOS. The STITCH-PELT variant uses the same penalty structure with the PELT solver \citep{Killick2011OptimalDO}.

STITCH-STD places a boundary at window $t$ when the adjacent cosine similarity falls below a data-adaptive threshold:
\[
\cos(\mathbf{e}_t, \mathbf{e}_{t+1}) < \mu_s - k\,\sigma_s,
\]
where $\mu_s$ and $\sigma_s$ are the mean and standard deviation of adjacent cosine similarities over the full window sequence, and $k = 2$ for all tasks.

Post-processing (Section~\ref{subsec:chunk_representation}) absorbs chunks shorter than a minimum duration into their neighbor. This threshold is $0.5\,$s for Kinetics-GEBD, $1.1\,$s for TAPOS, and $3.0\,$s for moment retrieval and VLM QA. The maximum chunk duration is $60\,$s universally. Parameter search was not extensive; values were set once per task on the available training split.

\paragraph{MMR frame selection.}
$\lambda$ and $\gamma$ are as defined in Section~\ref{sec:app:semantic}. For MMR chunk penalty we use $\lambda = 0.7$, $\gamma = 0.1$ for all three benchmarks (MLVU, LongVideoBench, VideoMME), selected via light search on the MLVU training split and applied to the other two without further tuning.
Chunk aggregation parameters for moment retrieval (GEM exponent $p$ and coherence temperature $\tau$) are noted in the captions of Tables~\ref{tab:moment_retrieval_anet} and~\ref{tab:moment_retrieval_qvh}.

\section{Compute Resources}
\label{sec:compute}

Our method is training-free and runs on a single consumer GPU (NVIDIA RTX 5080, 16\,GB VRAM, 32\,GB RAM), used primarily for InternVideo2 feature extraction. VLM inference for LLaVA-OV (7B) is served via a HuggingFace Inference Endpoint; Qwen3-VL-8B calls are made via the OpenRouter API. Both VLMs are queried with temperature $= 0$ and a maximum output length of 32 tokens, matching the format of the multiple-choice benchmarks (single-letter answer).

\section{Confidence Intervals}
\label{sec:ci}

Plots in Figure~\ref{fig:vlm_qa_plots} include 95\% confidence interval bands computed via non-parametric percentile bootstrap.
For each bin or frame-budget level, we treat each query as a Bernoulli outcome (1 if correct, 0 otherwise) and resample queries with replacement for 2000 iterations.
The lower and upper bounds of the band are the 2.5th and 97.5th percentiles of the resulting bootstrap distribution of the mean accuracy.
No normality assumption is made; the intervals are asymmetric by construction.
The intervals capture variability across queries within each bin, reflecting uncertainty due to finite query counts.

\section{Broader Impacts}
\label{sec:broader_impacts}

STITCH is a training-free method for converting long videos into reusable semantic chunks that can support event boundary detection, moment retrieval, and frame selection for video-language reasoning. The main positive impact of this work is improved efficiency and accessibility for long-video understanding: by reducing the need to process video densely or train task-specific models, STITCH may make video analysis more practical for researchers and practitioners with limited compute budgets. It may also support beneficial applications such as video search, assistive video navigation, educational content retrieval, and analysis of long recordings where compact temporal structure is useful.

At the same time, methods that make long videos easier to index, search, and summarize can also create risks. Improved temporal moment localization could be misused for surveillance, large-scale monitoring, or privacy-invasive search over personal or sensitive video collections. Incorrect temporal localization may be especially harmful in high-stakes settings, such as security, legal, medical, or workplace monitoring contexts, where a missed or misplaced event could affect downstream decisions. We therefore view STITCH as a general-purpose video analysis tool rather than a deployment-ready decision system. Appropriate use should include privacy safeguards and human oversight for high-stakes applications, as well as additional domain-specific evaluation before deployment.

%% file: main.bbl
\begin{thebibliography}{48}
\providecommand{\natexlab}[1]{#1}
\providecommand{\url}[1]{\texttt{#1}}
\expandafter\ifx\csname urlstyle\endcsname\relax
  \providecommand{\doi}[1]{doi: #1}\else
  \providecommand{\doi}{doi: \begingroup \urlstyle{rm}\Url}\fi

\bibitem[Afham et~al.(2023)Afham, Shukla, Poursaeed, Zhang, Shah, and
  Lim]{poursaeed2023revisitingkerneltemporal}
Mohamed Afham, Satya~Narayan Shukla, Omid Poursaeed, Pengchuan Zhang, Ashish
  Shah, and Sernam Lim.
\newblock Revisiting kernel temporal segmentation as an adaptive tokenizer for
  long-form video understanding.
\newblock In \emph{{ICCV} (Workshops)}, pages 1181--1186, 2023.
\newblock \doi{10.1109/ICCVW60793.2023.00128}.
\newblock URL \url{https://doi.org/10.1109/ICCVW60793.2023.00128}.

\bibitem[Bai et~al.(2025)Bai, Cai, Chen, Chen, Chen, Cheng, Deng, Ding, Gao,
  Ge, Ge, Guo, Huang, Huang, Huang, Hui, Jiang, Li, Li, Li, Li, Lin, Lin, Liu,
  Liu, Liu, Liu, Liu, Liu, Lu, Luo, Lv, Men, Meng, Ren, Ren, Song, Sun, Tang,
  Tu, Wan, Wang, Wang, Wang, Wang, Xie, Xu, Xu, Xu, Yang, Yang, Yang, Yang, Yu,
  Zhang, Zhang, Zhang, Zheng, Zhong, Zhou, Zhou, Zhou, Zhu, and Zhu]{Qwen3-VL}
Shuai Bai, Yuxuan Cai, Ruizhe Chen, Keqin Chen, Xionghui Chen, Zesen Cheng,
  Lianghao Deng, Wei Ding, Chang Gao, Chunjiang Ge, Wenbin Ge, Zhifang Guo,
  Qidong Huang, Jie Huang, Fei Huang, Binyuan Hui, Shutong Jiang, Zhaohai Li,
  Mingsheng Li, Mei Li, Kaixin Li, Zicheng Lin, Junyang Lin, Xuejing Liu,
  Jiawei Liu, Chenglong Liu, Yang Liu, Dayiheng Liu, Shixuan Liu, Dunjie Lu,
  Ruilin Luo, Chenxu Lv, Rui Men, Lingchen Meng, Xuancheng Ren, Xingzhang Ren,
  Sibo Song, Yuchong Sun, Jun Tang, Jianhong Tu, Jianqiang Wan, Peng Wang,
  Pengfei Wang, Qiuyue Wang, Yuxuan Wang, Tianbao Xie, Yiheng Xu, Haiyang Xu,
  Jin Xu, Zhibo Yang, Mingkun Yang, Jianxin Yang, An~Yang, Bowen Yu, Fei Zhang,
  Hang Zhang, Xi~Zhang, Bo~Zheng, Humen Zhong, Jingren Zhou, Fan Zhou, Jing
  Zhou, Yuanzhi Zhu, and Ke~Zhu.
\newblock Qwen3-vl technical report.
\newblock \emph{arXiv preprint arXiv:2511.21631}, 2025.

\bibitem[Bao and Kot(2024)]{bao2024simbase}
Peijun Bao and Alex~C. Kot.
\newblock Simbase: A simple baseline for temporal video grounding.
\newblock \emph{arXiv preprint arXiv:2411.07945}, 2024.

\bibitem[Carbonell and Goldstein(1998)]{carbonell1998mmr}
Jaime Carbonell and Jade Goldstein.
\newblock The use of mmr, diversity-based reranking for reordering documents
  and producing summaries.
\newblock In \emph{Proceedings of the 21st Annual International ACM SIGIR
  Conference on Research and Development in Information Retrieval}, SIGIR '98,
  page 335–336, New York, NY, USA, 1998. Association for Computing Machinery.
\newblock ISBN 1581130155.
\newblock \doi{10.1145/290941.291025}.
\newblock URL \url{https://doi.org/10.1145/290941.291025}.

\bibitem[Chasmai et~al.(2025)Chasmai, Jagatap, KV, Horn, Maji, and
  Fanelli]{Chasmai2025MomentSampling}
Mustafa Chasmai, Gauri Jagatap, Gouthaman KV, Grant~Van Horn, Subhransu Maji,
  and Andrea Fanelli.
\newblock Moment sampling in video llms for long-form video qa, 2025.
\newblock URL \url{https://arxiv.org/abs/2507.00033}.

\bibitem[Chen et~al.(2026)Chen, Zeng, Luo, Xie, Lin, Ji, Zhang, and
  Zheng]{chen2026wfsb}
Wang Chen, Yuhui Zeng, Yongdong Luo, Tianyu Xie, Luojun Lin, Jiayi Ji, Yan
  Zhang, and Xiawu Zheng.
\newblock Wavelet-based frame selection by detecting semantic boundary for long
  video understanding, 2026.
\newblock URL \url{https://arxiv.org/abs/2603.00512}.

\bibitem[Fang et~al.(2026)Fang, Song, Sun, Wu, Wu, and Chan]{fang2026threading}
Bo~Fang, YuXin Song, Haoyuan Sun, Qiangqiang Wu, Wenhao Wu, and Antoni~B. Chan.
\newblock Threading keyframe with narratives: {MLLM}s as strong long video
  comprehenders.
\newblock In \emph{The Fourteenth International Conference on Learning
  Representations}, 2026.
\newblock URL \url{https://openreview.net/forum?id=kyLS9EhPhY}.

\bibitem[Fu et~al.(2025)Fu, Dai, Luo, Li, Ren, Zhang, Wang, Zhou, Shen, Zhang,
  Chen, Li, Lin, Zhao, Li, Xu, Zheng, Chen, Shan, He, and Sun]{Fu2025VideoMME}
Chaoyou Fu, Yuhan Dai, Yongdong Luo, Lei Li, Shuhuai Ren, Renrui Zhang, Zihan
  Wang, Chenyu Zhou, Yunhang Shen, Mengdan Zhang, Peixian Chen, Yanwei Li,
  Shaohui Lin, Sirui Zhao, Ke~Li, Tong Xu, Xiawu Zheng, Enhong Chen, Caifeng
  Shan, Ran He, and Xing Sun.
\newblock Video-mme: The first-ever comprehensive evaluation benchmark of
  multi-modal llms in video analysis.
\newblock In \emph{2025 IEEE/CVF Conference on Computer Vision and Pattern
  Recognition (CVPR)}, pages 24108--24118, 2025.
\newblock \doi{10.1109/CVPR52734.2025.02245}.

\bibitem[Gordeev et~al.(2024)Gordeev, Dokholyan, Tolstykh, and
  Kuprashevich]{hou2024sgdetr}
Aleksandr Gordeev, Vladimir Dokholyan, Irina Tolstykh, and Maksim Kuprashevich.
\newblock Saliency-guided detr for moment retrieval and highlight detection,
  2024.
\newblock URL \url{https://arxiv.org/abs/2410.01615}.

\bibitem[Gothe et~al.(2024)Gothe, Agarwal, Ghosh, Vachhani, Kashyap, and
  Raja]{venkatesh2024flowgebd}
Sourabh~Vasant Gothe, Vibhav Agarwal, Sourav Ghosh, Jayesh~Rajkumar Vachhani,
  Pranay Kashyap, and Barath Raj~Kandur Raja.
\newblock What's in the flow? exploiting temporal motion cues for unsupervised
  generic event boundary detection.
\newblock \emph{CoRR}, abs/2404.18935, 2024.
\newblock URL \url{https://doi.org/10.48550/arXiv.2404.18935}.

\bibitem[Hwang et~al.(2025)Hwang, Gong, Kim, and Cho]{hwang2025diffgebd}
Jaejun Hwang, Dayoung Gong, Manjin Kim, and Minsu Cho.
\newblock Generic event boundary detection via denoising diffusion.
\newblock \emph{arXiv preprint arXiv:2508.12084}, 2025.

\bibitem[Jeon et~al.(2025)Jeon, Yang, Han, Hwang, Yoon, Kim, and
  Kim]{Jeon2025PointTS}
Mingyu Jeon, Jisoo Yang, Sungjin Han, Jinkwon Hwang, Sunjae Yoon, Jonghee Kim,
  and Junyeoung Kim.
\newblock Point to span: Zero-shot moment retrieval for navigating unseen
  hour-long videos.
\newblock \emph{ArXiv}, abs/2512.10363, 2025.
\newblock URL \url{https://api.semanticscholar.org/CorpusID:283737014}.

\bibitem[Killick et~al.(2012)Killick, Fearnhead, and
  Eckley]{Killick2011OptimalDO}
R.~Killick, P.~Fearnhead, and I.~A. Eckley.
\newblock Optimal detection of changepoints with a linear computational cost.
\newblock \emph{Journal of the American Statistical Association}, 107\penalty0
  (500):\penalty0 1590--1598, 2012.
\newblock \doi{10.1080/01621459.2012.737745}.
\newblock URL \url{https://doi.org/10.1080/01621459.2012.737745}.

\bibitem[Krishna et~al.(2017)Krishna, Hata, Ren, Fei-Fei, and
  Niebles]{krishna2017densecaption}
Ranjay Krishna, Kenji Hata, Frederic Ren, Li~Fei-Fei, and Juan~Carlos Niebles.
\newblock Dense-captioning events in videos.
\newblock In \emph{ArXiv}, 2017.

\bibitem[Lea et~al.(2016)Lea, Reiter, Vidal, and Hager]{lea2016segmental}
Colin Lea, Austin Reiter, Ren{\'e} Vidal, and Gregory~D. Hager.
\newblock Segmental spatiotemporal cnns for fine-grained action segmentation.
\newblock In Bastian Leibe, Jiri Matas, Nicu Sebe, and Max Welling, editors,
  \emph{Computer Vision -- ECCV 2016}, pages 36--52, Cham, 2016. Springer
  International Publishing.
\newblock ISBN 978-3-319-46487-9.

\bibitem[Lee et~al.(2025)Lee, Lee, Ahn, Choi, and Lee]{Lee_2025_BMVC}
Jin-Seop Lee, SungJoon Lee, Jaehan Ahn, YunSeok Choi, and Jee-Hyong Lee.
\newblock Tag: A simple yet effective temporal-aware approach for zero-shot
  video temporal grounding.
\newblock In \emph{36th British Machine Vision Conference 2025, {BMVC} 2025,
  Sheffield, UK, November 24-27, 2025}. BMVA, 2025.
\newblock URL
  \url{https://bmva-archive.org.uk/bmvc/2025/assets/papers/Paper_786/paper.pdf}.

\bibitem[Lei et~al.(2021)Lei, Berg, and Bansal]{Lei2021QVHighlightsDM}
Jie Lei, Tamara~Lee Berg, and Mohit Bansal.
\newblock Detecting moments and highlights in videos via natural language
  queries.
\newblock In A.~Beygelzimer, Y.~Dauphin, P.~Liang, and J.~Wortman Vaughan,
  editors, \emph{Advances in Neural Information Processing Systems}, 2021.
\newblock URL \url{https://openreview.net/forum?id=tfBBt_q4nHT}.

\bibitem[Li et~al.(2024)Li, Zhang, Zhang, Pu, Du, Dong, Liu, Zhang, Zhang, Li,
  and Liu]{lmms_eval2024}
Bo~Li, Peiyuan Zhang, Kaichen Zhang, Fanyi Pu, Xinrun Du, Yuhao Dong, Haotian
  Liu, Yuanhan Zhang, Ge~Zhang, Chunyuan Li, and Ziwei Liu.
\newblock Lmms-eval: Accelerating the development of large multimodal models,
  March 2024.
\newblock URL \url{https://github.com/EvolvingLMMs-Lab/lmms-eval}.

\bibitem[Li et~al.(2025)Li, Zhang, Guo, Zhang, Li, Zhang, Zhang, Zhang, Li,
  Liu, and Li]{li2025llavaonevision}
Bo~Li, Yuanhan Zhang, Dong Guo, Renrui Zhang, Feng Li, Hao Zhang, Kaichen
  Zhang, Peiyuan Zhang, Yanwei Li, Ziwei Liu, and Chunyuan Li.
\newblock {LL}a{VA}-onevision: Easy visual task transfer.
\newblock \emph{Transactions on Machine Learning Research}, 2025.
\newblock ISSN 2835-8856.
\newblock URL \url{https://openreview.net/forum?id=zKv8qULV6n}.

\bibitem[Li et~al.(2026)Li, Li, Tao, ZHAO, Wu, Zhao, Song, Niu, and
  Fazli]{li2026frameoracle}
Chaoyu Li, Tianzhi Li, Fei Tao, ZHENYU ZHAO, Ziqian Wu, Maozheng Zhao, Juntong
  Song, Cheng Niu, and Pooyan Fazli.
\newblock Frameoracle: Learning what to see and how much to see in videos,
  2026.
\newblock URL \url{https://openreview.net/forum?id=yBAp76dQMf}.

\bibitem[Lin et~al.(2023)Lin, Zhang, Chen, Pramanick, Gao, Wang, Yan, and
  Shou]{Lin2023UniVTG}
Kevin~Qinghong Lin, Pengchuan Zhang, Joya Chen, Shraman Pramanick, Difei Gao,
  Alex~Jinpeng Wang, Rui Yan, and Mike~Zheng Shou.
\newblock Univtg: Towards unified video-language temporal grounding.
\newblock In \emph{2023 IEEE/CVF International Conference on Computer Vision
  (ICCV)}, pages 2782--2792, 2023.
\newblock \doi{10.1109/ICCV51070.2023.00262}.

\bibitem[Lin et~al.(2019)Lin, Liu, Li, Ding, and Wen]{lin2019bmn}
Tianwei Lin, Xiao Liu, Xin Li, Errui Ding, and Shilei Wen.
\newblock {BMN}: Boundary-matching network for temporal action proposal
  generation.
\newblock In \emph{Proceedings of the {IEEE}/CVF International Conference on
  Computer Vision (ICCV)}, 2019.

\bibitem[Liu et~al.(2025)Liu, Zhao, Xu, and Ghanem]{liu2025bolt}
Shuming Liu, Chen Zhao, Tianqi Xu, and Bernard Ghanem.
\newblock Bolt: Boost large vision-language model without training for
  long-form video understanding.
\newblock In \emph{Proceedings of the IEEE/CVF Conference on Computer Vision
  and Pattern Recognition (CVPR)}, June 2025.

\bibitem[Luo et~al.(2024)Luo, Huang, Gong, Jin, and Liu]{Luo2024FrozenVLMVMR}
Dezhao Luo, Jiabo Huang, Shaogang Gong, Hailin Jin, and Yang Liu.
\newblock Zero-shot video moment retrieval from frozen vision-language models.
\newblock In \emph{2024 IEEE/CVF Winter Conference on Applications of Computer
  Vision (WACV)}, pages 5452--5461, 2024.
\newblock \doi{10.1109/WACV57701.2024.00538}.

\bibitem[Ma et~al.(2022)Ma, Xu, Sun, Yan, Zhang, and Ji]{ma2022xclip}
Yiwei Ma, Guohai Xu, Xiaoshuai Sun, Ming Yan, Ji~Zhang, and Rongrong Ji.
\newblock {X-CLIP:} end-to-end multi-grained contrastive learning for
  video-text retrieval.
\newblock \emph{arXiv preprint arXiv:2207.07285}, 2022.

\bibitem[Moon et~al.(2023)Moon, Hyun, Park, Park, and Heo]{Moon_2023_CVPR}
WonJun Moon, Sangeek Hyun, SangUk Park, Dongchan Park, and Jae-Pil Heo.
\newblock Query-dependent video representation for moment retrieval and
  highlight detection.
\newblock In \emph{Proceedings of the IEEE/CVF Conference on Computer Vision
  and Pattern Recognition}, pages 23023--23033, 2023.

\bibitem[Qu et~al.(2024)Qu, Chen, Liu, Li, and Zhao]{qu2024chatvtg}
Mengxue Qu, Xiaodong Chen, Wu~Liu, Alicia Li, and Yao Zhao.
\newblock { ChatVTG: Video Temporal Grounding via Chat with Video Dialogue
  Large Language Models }.
\newblock In \emph{2024 IEEE/CVF Conference on Computer Vision and Pattern
  Recognition Workshops (CVPRW)}, pages 1847--1856, Los Alamitos, CA, USA, June
  2024. IEEE Computer Society.
\newblock \doi{10.1109/CVPRW63382.2024.00191}.
\newblock URL
  \url{https://doi.ieeecomputersociety.org/10.1109/CVPRW63382.2024.00191}.

\bibitem[Radenović et~al.(2017)Radenović, Tolias, and Chum]{RadenovicTC19}
Filip Radenović, Giorgos Tolias, and Ondřej Chum.
\newblock Fine-tuning cnn image retrieval with no human annotation.
\newblock \emph{IEEE Transactions on Pattern Analysis and Machine
  Intelligence}, PP, 11 2017.
\newblock \doi{10.1109/TPAMI.2018.2846566}.

\bibitem[Radford et~al.(2021)Radford, Kim, Hallacy, Ramesh, Goh, Agarwal,
  Sastry, Askell, Mishkin, Clark, Krueger, and Sutskever]{radford2021clip}
Alec Radford, Jong~Wook Kim, Chris Hallacy, Aditya Ramesh, Gabriel Goh,
  Sandhini Agarwal, Girish Sastry, Amanda Askell, Pamela Mishkin, Jack Clark,
  Gretchen Krueger, and Ilya Sutskever.
\newblock Learning transferable visual models from natural language
  supervision.
\newblock In Marina Meila and Tong Zhang, editors, \emph{Proceedings of the
  38th International Conference on Machine Learning}, volume 139 of
  \emph{Proceedings of Machine Learning Research}, pages 8748--8763. PMLR,
  18--24 Jul 2021.
\newblock URL \url{https://proceedings.mlr.press/v139/radford21a.html}.

\bibitem[Shao et~al.(2020)Shao, Zhao, Dai, and Lin]{shao2020tapos}
Dian Shao, Yue Zhao, Bo~Dai, and Dahua Lin.
\newblock Intra- and inter-action understanding via temporal action parsing.
\newblock In \emph{IEEE Conference on Computer Vision and Pattern Recognition
  (CVPR)}, 2020.

\bibitem[Shou et~al.(2021)Shou, Lei, Wang, Ghadiyaram, and
  Feiszli]{shou2021gebd}
Mike~Zheng Shou, Stan~Weixian Lei, Weiyao Wang, Deepti Ghadiyaram, and Matt
  Feiszli.
\newblock Generic event boundary detection: A benchmark for event segmentation.
\newblock In \emph{Proceedings of the IEEE/CVF International Conference on
  Computer Vision (ICCV)}, pages 8075--8084, October 2021.

\bibitem[Sun et~al.(2024)Sun, Zhou, Chen, and Xie]{Sun_2024_AAAI_TRDETR}
Hao Sun, Mingyao Zhou, Wenjing Chen, and Wei Xie.
\newblock Tr-detr: Task-reciprocal transformer for joint moment retrieval and
  highlight detection.
\newblock In \emph{Proceedings of the AAAI Conference on Artificial
  Intelligence}, volume~38, pages 4998--5007, 2024.

\bibitem[Tan et~al.(2023)Tan, Wang, Wu, and Wang]{tan2023temporalperceiver}
Jing Tan, Yuhong Wang, Gangshan Wu, and Limin Wang.
\newblock Temporal perceiver: A general architecture for arbitrary boundary
  detection.
\newblock \emph{IEEE Transactions on Pattern Analysis and Machine
  Intelligence}, pages 1--16, 2023.
\newblock \doi{10.1109/TPAMI.2023.3283067}.

\bibitem[Tang et~al.(2025)Tang, Qiu, Xie, Tian, Jiao, and Ye]{tang2025adaptive}
Xi~Tang, Jihao Qiu, Lingxi Xie, Yunjie Tian, Jianbin Jiao, and Qixiang Ye.
\newblock Adaptive keyframe sampling for long video understanding.
\newblock \emph{arXiv preprint arXiv:2502.21271}, 2025.

\bibitem[Truong et~al.(2020)Truong, Oudre, and
  Vayatis]{Truong2020CPDImplementation}
Charles Truong, Laurent Oudre, and Nicolas Vayatis.
\newblock Selective review of offline change point detection methods.
\newblock \emph{Signal Processing}, 167:\penalty0 107299, 2020.
\newblock ISSN 0165-1684.
\newblock \doi{https://doi.org/10.1016/j.sigpro.2019.107299}.
\newblock URL
  \url{https://www.sciencedirect.com/science/article/pii/S0165168419303494}.

\bibitem[Wang et~al.(2024)Wang, Li, Li, Yu, He, Chen, Pei, Zheng, Wang, Shi,
  Jiang, Li, Xu, Zhang, Huang, Qiao, Wang, and Wang]{Wang2024InternVideo2SF}
Yi~Wang, Kunchang Li, Xinhao Li, Jiashuo Yu, Yinan He, Guo Chen, Baoqi Pei,
  Rongkun Zheng, Zun Wang, Yansong Shi, Tianxiang Jiang, Songze Li, Jilan Xu,
  Hongjie Zhang, Yifei Huang, Yu~Qiao, Yali Wang, and Limin Wang.
\newblock {InternVideo2}: Scaling foundation models for~multimodal video
  understanding.
\newblock In \emph{Computer Vision - {ECCV} 2024: 18th European Conference,
  Milan, Italy, September 29-October 4, 2024, Proceedings, Part {LXXXV}}, pages
  396--416, Berlin, Heidelberg, 2024. Springer-Verlag.
\newblock ISBN 978-3-031-73012-2.
\newblock \doi{10.1007/978-3-031-73013-9_23}.
\newblock URL \url{https://doi.org/10.1007/978-3-031-73013-9_23}.

\bibitem[Wang et~al.(2025)Wang, Yu, Stengel-Eskin, Yoon, Cheng, Bertasius, and
  Bansal]{Wang2024VideoTree}
Ziyang Wang, Shoubin Yu, Elias Stengel-Eskin, Jaehong Yoon, Feng Cheng, Gedas
  Bertasius, and Mohit Bansal.
\newblock Videotree: Adaptive tree-based video representation for llm reasoning
  on long videos.
\newblock In \emph{2025 IEEE/CVF Conference on Computer Vision and Pattern
  Recognition (CVPR)}, pages 3272--3282, 2025.
\newblock \doi{10.1109/CVPR52734.2025.00311}.

\bibitem[Wu et~al.(2024)Wu, Li, Chen, and Li]{wu2024longvideobench}
Haoning Wu, Dongxu Li, Bei Chen, and Junnan Li.
\newblock Longvideobench: A benchmark for long-context interleaved
  video-language understanding.
\newblock In \emph{The Thirty-eight Conference on Neural Information Processing
  Systems Datasets and Benchmarks Track}, 2024.
\newblock URL \url{https://openreview.net/forum?id=3G1ZDXOI4f}.

\bibitem[Xu et~al.(2025)Xu, Sun, Zhai, Li, Liang, Li, and
  Du]{xu2025zeroshot_vmr_mllm}
Yifang Xu, Yunzhuo Sun, Benxiang Zhai, Ming Li, Wenxin Liang, Yang Li, and
  Sidan Du.
\newblock Zero-shot video moment retrieval via off-the-shelf multimodal large
  language models.
\newblock In \emph{Proceedings of the Thirty-Ninth AAAI Conference on
  Artificial Intelligence and Thirty-Seventh Conference on Innovative
  Applications of Artificial Intelligence and Fifteenth Symposium on
  Educational Advances in Artificial Intelligence}, AAAI'25/IAAI'25/EAAI'25.
  AAAI Press, 2025.
\newblock ISBN 978-1-57735-897-8.
\newblock \doi{10.1609/aaai.v39i9.32971}.
\newblock URL \url{https://doi.org/10.1609/aaai.v39i9.32971}.

\bibitem[Yan et~al.(2023)Yan, Xiong, Nagrani, Arnab, Wang, Ge, Ross, and
  Schmid]{Yan2023UnLoc}
Shen Yan, Xuehan Xiong, Arsha Nagrani, Anurag Arnab, Zhonghao Wang, Weina Ge,
  David Ross, and Cordelia Schmid.
\newblock Unloc: A unified framework for video localization tasks.
\newblock In \emph{2023 IEEE/CVF International Conference on Computer Vision
  (ICCV)}, pages 13577--13587, 2023.
\newblock \doi{10.1109/ICCV51070.2023.01253}.

\bibitem[Yu et~al.(2025)Yu, JIN, Wang, Chen, Jin, ZUO, XIAOLEI, Sun, Zhang, Wu,
  Zhang, and Sun]{yu2025framevoyager}
Sicheng Yu, CHENGKAI JIN, Huanyu Wang, Zhenghao Chen, Sheng Jin, ZHONGRONG ZUO,
  XU~XIAOLEI, Zhenbang Sun, Bingni Zhang, Jiawei Wu, Hao Zhang, and Qianru Sun.
\newblock Frame-voyager: Learning to query frames for video large language
  models.
\newblock In \emph{The Thirteenth International Conference on Learning
  Representations}, 2025.
\newblock URL \url{https://openreview.net/forum?id=LNL7zKvm7e}.

\bibitem[Yuan et~al.(2022)Yuan, Ma, Wang, Liu, and Zhu]{yuan2019scdm}
Yitian Yuan, Lin Ma, Jingwen Wang, Wei Liu, and Wenwu Zhu.
\newblock Semantic conditioned dynamic modulation for temporal sentence
  grounding in videos.
\newblock \emph{IEEE Transactions on Pattern Analysis and Machine
  Intelligence}, 44\penalty0 (5):\penalty0 2725--2741, 2022.
\newblock \doi{10.1109/TPAMI.2020.3038993}.

\bibitem[Zhang et~al.(2025)Zhang, Sui, Liu, Mu, Wang, and
  Ghanem]{Zhang2025TimeLoc}
Chen-Lin Zhang, Lin Sui, Shuming Liu, Fangzhou Mu, Zhangcheng Wang, and Bernard
  Ghanem.
\newblock Timeloc: A unified end-to-end framework for precise timestamp
  localization in long videos, 2025.
\newblock URL \url{https://arxiv.org/abs/2503.06526}.

\bibitem[Zhang et~al.(2024)Zhang, Li, Zhang, Pu, Cahyono, Hu, Liu, Zhang, Yang,
  Li, and Liu]{zhang2024lmmsevalrealitycheckevaluation}
Kaichen Zhang, Bo~Li, Peiyuan Zhang, Fanyi Pu, Joshua~Adrian Cahyono, Kairui
  Hu, Shuai Liu, Yuanhan Zhang, Jingkang Yang, Chunyuan Li, and Ziwei Liu.
\newblock Lmms-eval: Reality check on the evaluation of large multimodal
  models, 2024.
\newblock URL \url{https://arxiv.org/abs/2407.12772}.

\bibitem[Zhang et~al.(2020)Zhang, Peng, Fu, and Luo]{luo2020learning2dtemporal}
Songyang Zhang, Houwen Peng, Jianlong Fu, and Jiebo Luo.
\newblock Learning 2d temporal adjacent networks for moment localization with
  natural language.
\newblock In \emph{The IEEE Conference on Computer Vision and Pattern
  Recognition (AAAI)}, 2020.

\bibitem[Zheng et~al.(2024{\natexlab{a}})Zheng, Cai, Chen, Peng, and
  Liu]{zheng2024tfvtg}
Minghang Zheng, Xinhao Cai, Qingchao Chen, Yuxin Peng, and Yang Liu.
\newblock Training-free video temporal grounding using large-scale pre-trained
  models.
\newblock In \emph{Computer Vision - ECCV 2024: 18th European Conference,
  Milan, Italy, September 29-October 4, 2024, Proceedings, Part LXXXII}, pages
  20--37, Berlin, Heidelberg, 2024{\natexlab{a}}. Springer-Verlag.
\newblock ISBN 978-3-031-73006-1.
\newblock \doi{10.1007/978-3-031-73007-8_2}.
\newblock URL \url{https://doi.org/10.1007/978-3-031-73007-8_2}.

\bibitem[Zheng et~al.(2024{\natexlab{b}})Zheng, He, Yang, and
  Li]{zheng2024dybdet}
Ziwei Zheng, Lijun He, Le~Yang, and Fan Li.
\newblock Fine-grained dynamic network for generic event boundary detection.
\newblock In \emph{Computer Vision – ECCV 2024: 18th European Conference,
  Milan, Italy, September 29–October 4, 2024, Proceedings, Part XLIII}, page
  107–123, Berlin, Heidelberg, 2024{\natexlab{b}}. Springer-Verlag.
\newblock ISBN 978-3-031-72774-0.
\newblock \doi{10.1007/978-3-031-72775-7_7}.
\newblock URL \url{https://doi.org/10.1007/978-3-031-72775-7_7}.

\bibitem[Zhou et~al.(2024)Zhou, Shu, Zhao, Wu, Xiao, Yang, Xiong, Zhang, Huang,
  and Liu]{Zhou2024MLVUBM}
Junjie Zhou, Yan Shu, Bo~Zhao, Boya Wu, Shitao Xiao, Xi~Yang, Yongping Xiong,
  Bo~Zhang, Tiejun Huang, and Zheng Liu.
\newblock Mlvu: A comprehensive benchmark for multi-task long video
  understanding.
\newblock \emph{arXiv preprint arXiv:2406.04264}, 2024.

\end{thebibliography}
